\documentclass{article}

\usepackage[preprint]{neurips_2025}

\PassOptionsToPackage{options}{natbib}

\usepackage[utf8]{inputenc} 
\usepackage[T1]{fontenc}    
\usepackage{hyperref}       
\usepackage{url}            
\usepackage{booktabs}       
\usepackage{amsfonts}       
\usepackage{nicefrac}       
\usepackage{microtype}      
\usepackage{xcolor}         

\usepackage{cite}
\usepackage{amsmath,amssymb,amsfonts}
\usepackage{graphicx}
\usepackage{textcomp}
\usepackage{subfigure}
\usepackage{mathtools}
\usepackage{amsthm}
\usepackage{algpseudocode}
\usepackage{wrapfig}
\usepackage{footnote}
\usepackage{comment}
\usepackage{float}
\usepackage{enumitem}

\usepackage{algorithm2e,refcount}
\SetKwInput{KwInput}{Input}
\SetKwInput{KwOutput}{Output}
\SetKwComment{Comment}{/* }{ */}
\RestyleAlgo{ruled}

\newtheorem{theoremEnv}{Theorem}[section]
\newtheorem{propositionEnv}[theoremEnv]{Proposition}
\newtheorem{lemmaEnv}[theoremEnv]{Lemma}

\input{my_symbol.sty}
\input{souvik.sty}

\newcommand{\newcite}[1]{\citeauthor{#1} [\citeyear{#1}]}
\newcommand{\yrcite}[1]{[\citeyear{#1}]}

\title{Landmark-Based Node Representations for Shortest Path Distance Approximations in Random Graphs}

\author{%
  My Le\\
  Dept. of Applied Mathematics \& Statistics\\
  Johns Hopkins University\\
  Baltimore, MD 21211 \\
  \texttt{mle19@jh.edu} \\
  \And
  Luana Ruiz \\
  Dept. of Applied Mathematics \& Statistics \\
  Johns Hopkins University\\
  Baltimore, MD 21211 \\
  \texttt{lrubini1@jh.edu} \\
  \AND
  Souvik Dhara \\
  School of Industrial and Systems Engineering\\
  Georgia Institute of Technology \\
  Atlanta, GA 30332 \\
  \texttt{sdhara@gatech.edu} \\
}

\begin{document}

\maketitle

\begin{abstract}
  Learning node representations is a fundamental problem in graph machine learning. While existing embedding methods effectively preserve local similarity measures, they often fail to capture global functions like graph distances. Inspired by Bourgain's seminal work on Hilbert space embeddings of metric spaces \yrcite{Bourgain85}, we study the performance of local distance-preserving node embeddings. Known as landmark-based algorithms, these embeddings approximate pairwise distances by computing shortest paths from a small subset of reference nodes called landmarks. Our main theoretical contribution shows that random graphs, such as Erd\H{o}s--R\'enyi random graphs, require lower dimensions in landmark-based embeddings compared to worst-case graphs. Empirically, we demonstrate that the GNN-based approximations for the distances to landmarks generalize well to larger real-world networks, offering a scalable and transferable alternative for graph representation learning.
\end{abstract}

\section{Introduction}

\subsection{Motivations}

Learning representations for network data has long been central to graph machine learning with a key objective to learn low-dimensional \textit{node embeddings} that map structurally similar nodes to nearby points. These embeddings facilitate the application of machine learning methods to graph data, enabling a wide range of downstream tasks such as node classification, link prediction, and community detection \citep{hamilton2017representation, grover2016node2vec}.  

Traditional methods such as DeepWalk \citep{perozzi2014deepwalk} and Node2Vec \citep{grover2016node2vec} use random walks to preserve local graph structures like node neighborhoods, while extensions such as GraRep \citep{cao2015grarep} and PRONE \citep{zhang2021prone} capture higher-order relationships via $k$-hop transition factorizations. Spectral methods like Laplacian Eigenmaps \citep{belkin2003laplacian} preserve global geometry by embedding graphs into low-dimensional spaces that approximate their underlying manifold. Cauchy embeddings \citep{tang2019cauchy} further improve spectral methods by increasing their sensitivity to edge weight differences. While effective at capturing local graph structure, these methods often fail to preserve global topology and functionals such as shortest path distances, especially in large, complex graphs \citep{goyal2018graph, tsitsulin2018just, brunner2021distance}. 

In this work, we focus on the problem of learning node embeddings that preserve both local similarities and global graph distances. Motivated by Bourgain’s seminal results on metric embeddings \yrcite{Bourgain85}, we analyze a landmark-based algorithm that approximates graph distances via shortest paths from a small set of landmarks. Our study analyzes its performance on random graphs—particularly Erd\H{o}s--R\'enyi (ER) graphs—compared to worst-case instances. Of particular interest is the dimension of the embedding space.

\subsection{Our Contributions}

\noindent\textbf{Theoretical Contributions.} Our theoretical contribution is a detailed analysis of the dimensionality requirements for landmark-based embeddings on random graphs, in a more generalized setting than that analyzed for worst-case graphs. \emph{This is the primary contribution of our work.}

We show that, with high probability (w.h.p.), random graphs require lower embedding dimensions: $\Omega\left(n^{\frac{1}{2c-1}+\varsigma}\theta\log n\right)$ with $\theta\in[\frac{c - 1}{2c - 1},\frac{2(c - 1)}{2c - 1})$ for a $\frac{1}{2c-1}$-factor lower bound, and $\Omega(n^{3-2c+\varsigma})$ for a $(2c-1)$-factor upper bound for any $\varsigma>0$, as compared to worst-case graphs with $\Omega(n^{1/c}\log n)$ for the same lower and upper bounds \citep{Bourgain85,Mat96,Sarma2010ASD}, where $c>1$. The proof leverages branching process approximations from the random graph literature \citep{RGCN1,RGCN2}.

\noindent\textbf{Methodological Contributions.} Building upon this theory, we propose a GNN-augmented variant that predicts landmark distances from graph structure. This reduces explicit shortest-path computations as GNNs can learn to approximate landmark distances in a supervised manner.

GNNs are well-suited for this task due to their alignment with dynamic programming which underpins shortest path algorithms~\citep{xu2019can, dudzik2022graph}. Empirical results on ER graphs and real-world benchmarks show that GNN-based embeddings provide better global-distance lower bounds than exact landmark embeddings. Notably, GNNs trained on small ER graphs generalize effectively to larger ER graphs and real-world networks, highlighting the value of studying embedding methods in the context of random graphs.

\subsection{Related Works}

A rich body of theoretical works has focused on the minimum dimension $k_\varepsilon$ required to embed worst-case graphs into $\R^k$ while preserving all pairwise distances up to a factor of $(1\pm \varepsilon)$. In a seminal work, \newcite{Bourgain85} showed $k_\varepsilon = \Omega((\log n)^2/ (\log \log n)^2)$, providing a negative answer to Johnson and Lindenstrauss's Problem 3 \yrcite{JL84}. This was later strengthened to $k_\varepsilon=\Omega((\log n)^2)$ \citep{LLR95}, and further to $k_\varepsilon=\Omega(n^{c/(1+\varepsilon)})$ for some universal $c>0$ \citep{Mat96}. The latter was also proven recently by \newcite{Naor16} and \newcite{Naor21} using expanders, showing that low-distortion embeddings of graphs with strong expansion properties require polynomial dimensionality. 

From an algorithmic perspective, finding embeddings with minimum distortion is NP-hard; see \newcite{Sid19} for a survey of approximation algorithms and hardness results. Practical methods often rely on landmark-based algorithms \citep{goldberg2005computing,Sarma2010ASD,potamias2009fast,tretyakov2011fast,akiba2013fast,rizi2018shortest,Qi2020ALB}, which preselect a subset of landmark nodes and compute distances to them via local message passing (see \newcite{Sommer2014queries} for a review). The resulting landmark distances can be viewed as an embedding useful for approximating graph distances. Yet these methods inherit worst-case limitations of local message passing, often requiring prohibitively large dimensions for general graphs \citep{Sarma2012, Loukas2020}.

\noindent \textbf{Notation.}~We let $G=(V,E)$ denote an undirected, unweighted graph, where $V$ is the set of nodes and $E$ is the set of edges, with $|V| = n$ and $|E| = m$, where $|X|$ denotes the cardinality of any discrete set $X$. We consider only one graph at a time and use $\mathcal{C}_{\sss (i)}$ to denote the $i$-th largest connected component of the graph. We write $u_1\leftrightarrow u_2$ to mean that there exists a path between $u_1$ and $u_2$ (i.e., $u_1$ and $u_2$ are in the same connected component). We often use the Bachmann--Landau asymptotic notation $o(1), O(1), \omega(1), \Omega(1), \Theta(1)$, etc. to describe the asymptotic behavior of functions.
Given a sequence of probability measures $(\PR_n)_{n\geq 1}$, a sequence of events $(\cE_n)_{n\geq 1}$ is said to hold \emph{with high probability (w.h.p.)} if $\lim_{n\to\infty} \PR_n(\cE_n) = 1$. For a sequence of random variables $(X_n)_{n\geq 1}$, $X_n \xrightarrow{\sss\PR} c$ means that $X_n$ converges to $c$ in probability. We write statements such as $X_n = f(n)^{o(1)}$ w.h.p. to abbreviate that $\log X_n /\log f(n) \xrightarrow{\sss\PR} 0$. Also, we write $X_n = O(1)$ w.h.p. to mean that $\PR(X_n \geq K) \to 0$ for a sufficiently large $K$.

\section{The Shortest Path Problem and Landmark-Based Embeddings}\label{section2}

Given a graph \( G = (V, E) \) and nodes \( u_1, u_2 \in V \), the shortest distance problem is to find the minimum number of edges connecting \( u_1 \) and \( u_2 \), i.e., the \textit{shortest path distance} \( d(u_1,u_2) \). The classical solution to this fundamental graph problem is Dijkstra’s algorithm, with running time \( O(n^2) \) for a single pair and \( O(n^3) \) for all pairs using naive data structures, reducible to \( O(m \log n) \) and \( O(m+n \log n) \) with heaps and Fibonacci heaps \citep{schrijver2012history}. More refined variants for single source include S-Dial (\( O(m+n l_{\max}) \), $l_{\max}$ is the maximum arc length), S-Heap (\( O(m \log n) \) or \( O(n \log n) \) in sparse graphs), and Fredman–Willard’s implementation (\( O(m+n \log n /\log\log n) \)) \citep{gallo1988shortest, fredman1990trans}. For all pairs, Floyd–Warshall and primal sequential algorithms run in \( O(n^3) \) \citep{gallo1988shortest}, while hidden-path achieves $\smash{O(mn+n^2\log n)}$ \citep{karger1993finding}. Despite these advances, exact computation remains costly on large graphs.

While computing exact shortest path distances is expensive, we can afford to compute local paths. Sarma et al.'s offline sketch algorithm \yrcite{Sarma2010ASD} leverages this principle in its local step to construct landmark embeddings (see Local Step in Algorithm \ref{alg}). To mitigate the time and memory constraints associated with calculating shortest paths, lower and upper bounds have been used as reliable metrics for approximating shortest paths in many approaches \citep{Bourgain85,Mat96,Sarma2010ASD,Gubichev2010,Sommer2014queries,Akiba2014queries,Meng2015GRECS,Jiang2021Tripoline,Awasthi2021}. As in the current setting, the resulting local embeddings can be stored in memory and later retrieved to quickly estimate $d(u_1,u_2)$ via the bounds $\underline{d}(u_1,u_2)$ and $\bar{d}(u_1,u_2)$ with a single lookup from $u_1$ and $u_2$ (see Global Step in Algorithm \ref{alg}).

\begin{algorithm}[H]
\caption{Landmark Algorithm Adapted From \newcite{Sarma2010ASD}}\label{alg}
\KwInput{Connected graph $G=(V,E)$ with $|V|=n$. Positive integer $R$. Set cardinalities $|S_0|=1$, $|S_1|$, $|S_2|$, ..., $|S_r|$ for some positive integer $r$.}
\KwOutput{Shortest path lower bound $\underline{d}(u,v)$ and upper bound $\bar{d}(u,v)$ for all $u,v \in V$.}
\For{$i = 1,2, \ldots, R$ \qquad\qquad\qquad\qquad\qquad\qquad\qquad\qquad\qquad\qquad\qquad\tcc{LOCAL STEP}}{ 
\For{$j = 0,1, \ldots, r$}{ 
$S_j \gets \{s_1, \ldots, s_{|S_j|} \sim$ Uniform$(V)$\}

\For{$u = 1, \ldots, n$}{
$[\bbx_u]_j = \min_{s \in S_j}d(s,u)$

$[\boldsymbol{\sigma}_u]_j = \mbox{argmin}_{s \in S_j}d(s,u)$
}
}
}
\For{$u = 1, \ldots, n$ \qquad\qquad\qquad\qquad\qquad\qquad\qquad\qquad\qquad\qquad\qquad\:\tcc{GLOBAL STEP}}{
\For{$v = 1, \ldots, n$}{

$\underline{d}(u,v) = \|\bbx_u-\bbx_v\|_\infty$ \qquad\qquad\qquad\qquad\qquad\qquad\qquad\:\:\:\:\:\:\:\:\:\:\:\:\tcc{Lower Bound}

$\bar{d}(u,v) = \min\{[\bbx_{u}]_i+[\bbx_{v}]_j:(i,j) \text{ s.t. } [\boldsymbol{\sigma}_{u} \bold{1}^T-\bold{1}\boldsymbol{\sigma}_{v}^T]_{ij}=0\}$\:\:\:\:\:\tcc{Upper Bound}
}
}
\end{algorithm}

To show that $\underline{d}(u_1,u_2)$ is a lower bound on $d(u_1,u_2)$, without loss of generality, assume $\smash{\underline{d}(u_1,u_2)=d(u_1,S)-d(u_2,S)}$ for some landmark set $S$ with $\smash{d(u_2,S)=d(u_2,v_1)}$ and $\smash{d(u_1,S)=d(u_1,v_2)\leq d(u_1,v_1)}$. It then follows from triangle inequality that $\smash{\underline{d}(u_1,u_2)\leq d(u_1,v_1)-d(u_2,v_1)\leq d(u_1,u_2)}$.

The proof for $d(u_1,u_2)\leq \bar{d}(u_1,u_2)$ also follows directly from the formulation of $\bar{d}(u_1,u_2)$ in Algorithm \ref{alg} and triangle inequality: $\smash{\bar{d}(u_1,u_2)=d(u_1,v)+d(u_2,v)\geq d(u_1,u_2)}$ for some landmark node $v$. By sampling at least one landmark set of size~1, we ensure that $u_1$ and $u_2$ share a closest landmark node from such landmark sets, preventing $\bar{d}(u_1,u_2)$ from being undefined.

Since $\underline{d}(u_1,u_2)$ depends on the distance to the landmark sets but not on which landmark node is the closest, it is sufficient for the landmark embeddings to store only the closest distances from each node to the landmark sets. The trade-off for such memory reduction is that $\underline{d}(u_1,u_2)$ can be approximated only with $D = R\times (r+1)$ dimensions, while $\bar{d}(u_1,u_2)$, which utilizes the common closest landmarks, has an approximation dimension that varies between $1$ and $D \times D$, depending on how the landmark sets are sampled.

\section{Lower and Upper Bound Distortions for Shortest Distance Approximations} \label{sec:worstcase}

The lower and upper bound metrics on the landmark embeddings, as described in Section \ref{section2}, are only useful if we can derive guarantees on their approximation ability. For the lower bound, these have been proven by \newcite{Mat96} based on Bourgain's classical embedding theorem \yrcite{Bourgain85}, which characterizes the distortion incurred by optimal embeddings of metric spaces into $\reals^{D}$. For the upper bound, similar guarantees have been derived in \newcite{Sarma2010ASD}.

\begin{theoremEnv}[\textbf{Lower Bound Distortion Adapted From \newcite{Bourgain85} and \newcite{Mat96}}]\label{thm:bourgain_distortion}
Let $G$ be a graph with $n \geq 3$ nodes and $u_1,u_2$ be two nodes in $G$. Let $c > 1$. There exist node embeddings $\bbx^*_{u_1},\bbx^*_{u_2} \in \reals^{D}$ with $D = \Omega (n^{1/c}\log{n})$ for which $\underline{d}(u_1,u_2)$ as in Algorithm \ref{alg} satisfies
\begin{equation} \label{eqn:bourgain_distortion}
\frac{d(u_1,u_2)}{2c-1} \leq \underline{d}(u_1,u_2)\leq d(u_1,u_2).
\end{equation}
\end{theoremEnv}

\begin{theoremEnv}[\textbf{Upper Bound Distortion Adapted From \newcite{Sarma2010ASD}}]\label{thm:sarma_distortion}
Let $G$ be a graph with $n \geq 3$ nodes and $u_1,u_2$ be two nodes in $G$. Let $c > 1$. There exist node embeddings $\bbx^*_{u_1},\bbx^*_{u_2} \in \reals^{D}$ with $D = \Omega (n^{1/c}\log{n})$ for which $\bar{d}(u_1,u_2)$ as in Algorithm \ref{alg} satisfies
\begin{equation} \label{eqn:sarma_distortion}
{d(u_1,u_2)} \leq \bar{d}(u_1,u_2) \leq (2c-1)d(u_1,u_2) \text{.}
\end{equation}
\end{theoremEnv}

In order for \eqref{eqn:bourgain_distortion} and \eqref{eqn:sarma_distortion} to hold, the embeddings $\bbx^*_u$ need to be optimal. However, there is no guarantee that this can be achieved using the landmark embeddings. One way to ensure good embeddings is to control how the landmarks are sampled. \newcite{Sarma2010ASD} proposed sampling landmark sets $S_i$ of sizes $2^i$ for $i = 0, 1, \ldots, r$.

For the lower bound, smaller landmark sets are beneficial since, for $\sigma_1 + \sigma_2 < 1$ with $0 < \sigma_1 < \sigma_2$, we must find at least one landmark set containing a landmark node in the $\sigma_1 d(u_1,u_2)$-hop neighborhood centered at $u_1$ and none in the $\sigma_2 d(u_1,u_2)$-hop neighborhood centered at $u_2$. For the upper bound, this strategy ensures that a landmark falls in the intersection of the $\lceil \tfrac{d(u_1,u_2)}{2} \rceil$-hop neighborhoods of nodes $u_1$ and $u_2$ w.h.p. Hence, having a range of landmark set cardinalities helps.

It can be shown that if $|S_i|$ is exponential in $i$, $R = \Omega(n^{1/c})$, and $r = \lfloor \log n \rfloor$---yielding a total embedding size of $\Theta(n^{1/c}\log n)$---then the resulting shortest path distance approximations satisfy Theorems \ref{thm:bourgain_distortion} and \ref{thm:sarma_distortion} for all pairs of nodes w.h.p. for \emph{any graph}. In Section \ref{sec:theory}, we show that both the distortions and the embedding dimensions can be improved for random graphs.

\section{Lower and Upper Bound Distortions on Sparse Erd\H{o}s--R\'enyi} \label{sec:theory}

In this section, we show our main results on the performance of $\underline{d}(u_1,u_2)$ and $\bar{d}(u_1,u_2)$ outputted by Algorithm \ref{alg} as shortest path approximations in sparse ER graphs, where each edge appears independently with a fixed probability. We write $G \sim \ER(\lambda/n)$ to denote this distribution over the space of all graphs on $n$ nodes with probability $\lambda/n$, $\lambda\in [0,n]$. Based on a classical result in random graph theory \citep[Theorems~4.4,~4.8 and Corollary~4.13]{RGCN1}, we consider $\lambda>1$ since otherwise the giant component dies out in probability, making most pairs of nodes not connected.

\subsection{Main Results on Distortions}

On ER graphs, we derive the following distortion bound as a $(1 \pm \varepsilon)$-approximation of $d(u_1,u_2)$:

\begin{theoremEnv}[\textbf{\textit{Lower Bound Distortion on Random Graphs}}]\label{thm:main_lb}
  Let $G\sim \ER(\lambda/n)$ with $\lambda>1$. Let $u_1,u_2$ be chosen independently and uniformly at random with replacement from $G$. Fix $\smash{\varepsilon\in (0,1)}$, an integer $M>1$, $\theta\in (0,\varepsilon)$, and $r=\lfloor \tfrac{\theta}{\log M}\log n \rfloor$. With embedding dimension $\smash{D=\Omega \left(M n^{1-\frac{\varepsilon}{2}-\min\left\{\frac{\varepsilon}{2},\theta\right\}+\varsigma}\tfrac{\theta}{\log M}\log n\right)}$ resulting from $R=\Omega\left(M n^{1-\frac{\varepsilon}{2}-\min\left\{\frac{\varepsilon}{2},\theta\right\}+\varsigma}\right)$ runs of the local step with set cardinalities $|S_0|=M^0, |S_1|=M^1, \ldots, |S_r|=M^r$ and any arbitrarily small $\varsigma>0$, $\underline{d}(u_1,u_2)$ provides a $(1-\varepsilon)$-approximation of $d(u_1,u_2)$ (i.e. $\smash{d(u_1,u_2)\geq \underline{d}(u_1,u_2)\geq (1-\varepsilon) d(u_1,u_2)}$) w.h.p.
\end{theoremEnv}

\begin{theoremEnv}[\textbf{\textit{Upper Bound Distortion on Random Graphs}}]\label{thm:main_ub}
  Let $G,\lambda,u_1,u_2$ be as in Theorem \ref{thm:main_lb}. Fix $\varepsilon\in (0,1)$, an integer $M>1$, $\theta\in \left(0,\frac{1-\varepsilon}{2}\right)$, and $r=\lfloor \tfrac{\theta}{\log M}\log n \rfloor$. With embedding dimension $\smash{D=\Omega \left(n^{1-\varepsilon+\varsigma}\right)}$ resulting from $R=\Omega\left(\frac{\log M}{\theta\log n}n^{1-\varepsilon+\varsigma}\right)$ runs of the local step with set cardinalities $|S_0|=M^0, \smash{|S_1|=M^1}, \ldots, |S_r|=M^r$ and any arbitrarily small $\varsigma>0$, $\bar{d}(u_1,u_2)$ provides a $(1+\varepsilon)$-approximation of $d(u_1,u_2)$ (i.e. $d(u_1,u_2)\leq \bar{d}(u_1,u_2)\leq (1+\varepsilon) d(u_1,u_2)$) w.h.p.
\end{theoremEnv}

While \newcite{Bourgain85}, \newcite{Mat96}, and \newcite{Sarma2010ASD} showed that, in the worst case, Algorithm \ref{alg} with $M=2$ requires embedding dimension $\Omega(n^{1/c}\log n)$ for a $\frac{1}{2c-1}$-factor lower bound and a $(2c-1)$-factor upper bound ($c>1$), our results offer a more efficient alternative for ER graphs by loosening the dimensionality restrictions, specifically $\Omega\left(n^{\frac{1}{2c-1}+\varsigma}\theta\log n\right)$ with $\theta\in[\frac{\varepsilon}{2},\varepsilon)$ for the same lower bound and $\Omega(n^{3-2c+\varsigma})$ for the same upper bound for any $\varsigma>0$. Furthermore, our results pertain to a more general setting where $M$ can be any integer greater than 1 and $\theta$, which regulates the amount of sampling, can be $\varepsilon$-small for the lower bound distortion and $\left(\tfrac{1-\varepsilon}{2}\right)$-small for the upper bound distortion.

\subsection{Idea of Proofs and Supporting Results}

The proofs of Theorems~\ref{thm:main_lb} and~\ref{thm:main_ub} rely on local neighborhood expansions in ER graphs $\smash{G\sim \ER(\lambda/n)}$, which can be accessed via the Poisson branching process with mean offspring $\lambda$. With $N_k(u)$ denoting the set of nodes with graph distance at most $k$ from $u$ and $\partial N_k(u)$ denoting the set of nodes at distance exactly $k$ from $u$, the results on local neighborhood expansions are stated as follows:

\begin{lemmaEnv}\label{lem:event-equivalence}
Let $G,\lambda,u_1,u_2$ be as in Theorems \ref{thm:main_lb} and \ref{thm:main_ub}. Let $\kappa_0\in (0,\frac{1}{2})$, $L = \kappa_0 \log_{\lambda} n$, and $\varepsilon \in (0,\kappa_0)$. 
Let $A_n$ denote the event that $n^{-\varepsilon}\lambda^L \leq |\partial N_L(u_1)|,|\partial N_L(u_2)| \leq n^{\varepsilon}\lambda^L$ and $B_n $ denote the event that $u_1$ and $u_
2$ are in the same connected component. Then $\PR(A_n \setminus B_n) \to 0$ and $\PR(B_n \setminus A_n) \to 0$ as $n\to\infty$.
\end{lemmaEnv}

\begin{proof}
    See Appendix \ref{app:pf3.5}.
\end{proof}

\begin{lemmaEnv}\label{lem:concentration-neighbohoods}
Let $G,\lambda,u_1,u_2,k_0,L$ be as in Lemma \ref{lem:event-equivalence}. Let $\varepsilon \in (0,\kappa_0)$ and $\kappa\in (0,1-\kappa_0)$.
Let $A_{b_m,b_M}$ be the event that $|\partial N_L(u_i)| \in [b_m,b_M]$ for $i\in \{1,2\}$ and $\cE_{n,k}$ be the good event that $|\partial N_{L+k}(u_i)| \in [b_m \lambda^{k}, b_M \lambda^{k}]$ for $i\in \{1,2\}$, where $b_m=n^{-\varepsilon} \lambda^L$ and $b_M=n^{\varepsilon} \lambda^L$.
Then, there exists $\delta>0$ such that $\PR(\cap_{l=0}^k\cE_{n,l} \mid A_{b_m,b_M}) \geq 1-3kn^{-\delta}$ for any $k\leq \kappa\log_{\lambda}n$ for all sufficiently large $n$.
\end{lemmaEnv}

\begin{proof}
    See Appendix \ref{app:pf3.4}
\end{proof}

\begin{propositionEnv} \label{prop:union-intersection}
  Let $G,\lambda,u_1,u_2,\kappa_0,\kappa, L$ be as in Lemma \ref{lem:concentration-neighbohoods} and $\varepsilon>0$. Conditionally on $u_1, u_2$ being in the same connected component, $|\partial N_{k_1}(u_1) \cap \partial N_{k_2}(u_2)|\in \left[\frac{n^{-\varepsilon}\lambda^{k_1+k_2}}{2n}, \frac{n^{\varepsilon}\lambda^{k_1+k_2}}{n}\right]$ w.h.p. for any $k_1, k_2$ such that $L< k_1, k_2\leq (\kappa_0+\kappa)\log_{\lambda} n$ and $k_1 + k_2 > (1 + \zeta) \log_{\lambda} n$ for any small $\zeta > 0$.
\end{propositionEnv}

\begin{proof}
    See Appendix \ref{app_prop}
\end{proof}

By Lemmas~\ref{lem:event-equivalence} and~\ref{lem:concentration-neighbohoods}, $\partial N_k(u_1)$ grows as $\lambda^k$ for any fixed $u$. By Proposition~\ref{prop:union-intersection}, $\smash{|\partial N_k(u_1) \cap \partial N_k(u_2)|}$ grows as $\tfrac{\lambda^{2k}}{n}$ for any fixed $u_1,u_2$. These growth rates imply that, w.h.p., the local step selects a landmark set that intersects $N_{k_1}(u_1)$ but not the disjoint $N_{k_2}(u_2)$, with $k_2 - k_1 \geq (1-\varepsilon)d(u_1,u_2)$, yielding \[\underline{d}(u_1,u_2) \:\geq\: (1-\varepsilon)\,d(u_1,u_2) \text { w.h.p.}\]
For the upper bound distortion, we show that w.h.p. there is a landmark set intersecting $\smash{N_k(u_1) \cap N_k(u_2)}$ but not $\bigl(N_k(u_1) \cup N_k(u_2)\bigr)\setminus\bigl(N_k(u_1)\cap N_k(u_2)\bigr)$, where $k \leq \tfrac{1+\varepsilon}{2}d(u_1,u_2)$. This ensures \[\overline{d}(u_1,u_2) \:\leq\: (1+\varepsilon)\,d(u_1,u_2) \text { w.h.p.}\] The complete proof of Theorems~\ref{thm:main_lb} and~\ref{thm:main_ub} are provided in Appendices \ref{app_lb} and \ref{app_ub}.

\section{GNN-Based Landmark Embeddings and Experimental Results} \label{sec:numerical}

Although Algorithm \ref{alg} outperforms traditional methods, its landmark distance calculations rely on one run of Breadth-First Search (BFS) for each landmark set, which is costly for large graphs (\(O(n+m)\) per pair, \(O(n(n+m))\) for all pairs \citep{cormen2009introduction}). We propose replacing BFS with a GNN to approximate shortest-path distances between nodes and landmarks, which comes with three advantages: (i) embeddings are computed automatically once the GNN is trained, (ii) inference is cheaper than exact distance calculations, and (iii) the GNN’s transferability \citep{ruiz20-transf,ruiz2021transferability} enables generalization to larger graphs from the same graphon model. 

\begin{figure*}[h]
\centering
\includegraphics[width=1\textwidth]{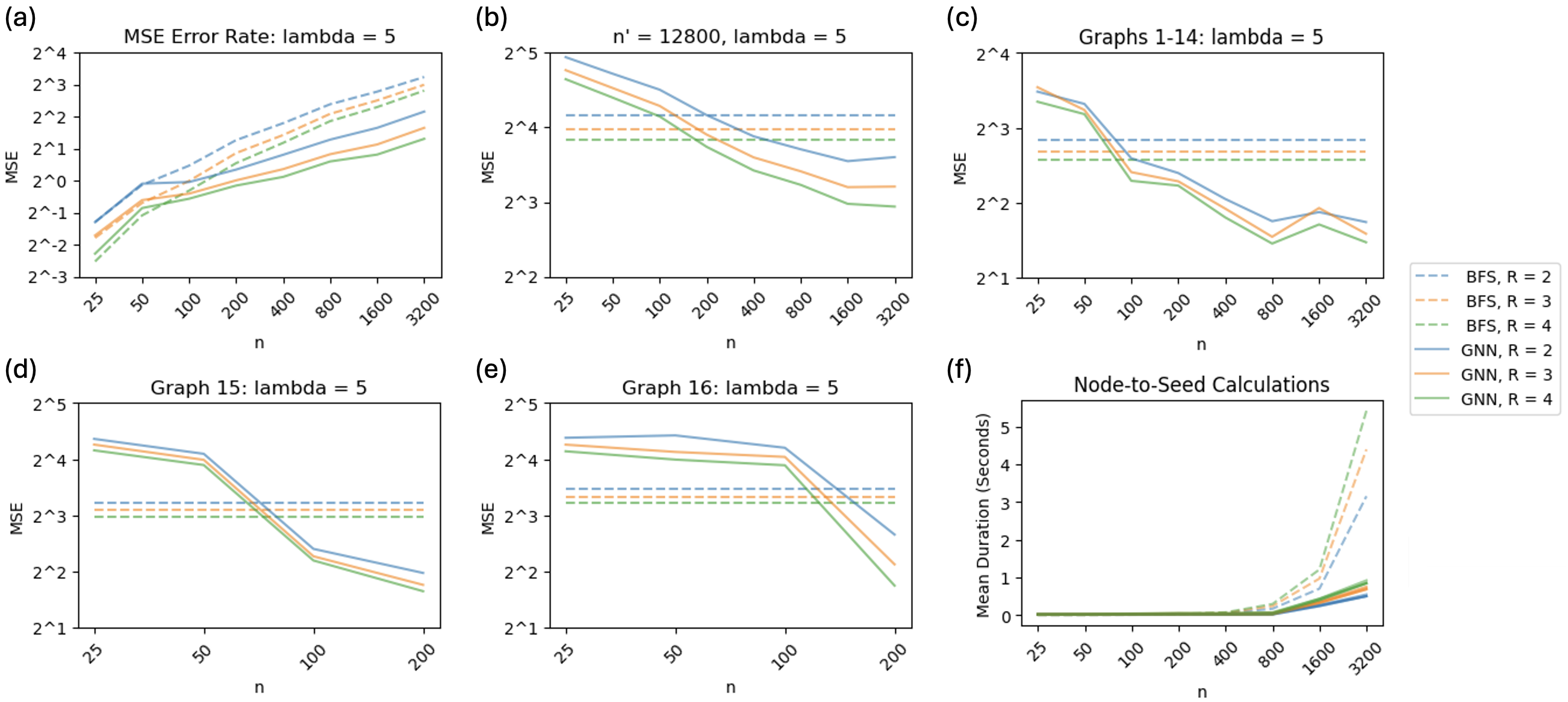}
\caption{Error rates of BFS-based and GNN-based lower bounds on (a) test ER graphs generated from the same $\ER(\lambda/n)$ as the training graphs, (b) test ER graphs generated by $\text{ER}_{n'}(\lambda/n')$ with larger graph size $n'$, (c) real-world networks with 3,892 to 28,281 nodes, (d) Brightkite social network with 56,739 nodes, and (e) ER-AVGDEG10-100K-L2 labeled network with 99,997 nodes. (f) Duration of generating all landmark distances by NetworkX's highly optimized BFS compared with our widest and deepest GNNs---GCN, GraphSage, GAT, and GIN models were examined and are represented by solid lines of the same color for the same number of local step $R$. See Appendices \ref{app:exp_details} and \ref{app:more_exps} for further details and discussions on the experiments and benchmark networks.}
\label{fig1}
\end{figure*}

GNNs are well-suited for this task as they align with dynamic programming strategies that are used in shortest-path algorithms \citep{xu2019can, dudzik2022graph}. As shown in Figure \ref{fig1}(a), the GNN achieves a substantial improvement over the vanilla lower bounds in approximating shortest path distances across all tested $R$, aided by better-learned embeddings and the near-certain connectivity of large graphs in this regime. Figures \ref{fig1}(b-e) further demonstrate the transferability of GNNs to larger ER graphs and real-world benchmark datasets. Particularly, the GNN-based embeddings achieve comparable or better performance than BFS-based embeddings, with MSE steadily improving as training graph size increases, even when learned on synthesized graphs up to 128 times smaller than the target graph. The GNN-based embeddings not only provide better distance approximations but also scale more efficiently in time and space than their BFS-based counterparts, as illustrated in Figure~\ref{fig1}(f), making them a promising tool for large-scale graph representation learning in practical applications.

\section{Conclusion}
\label{others}

Our analysis, focused on average-case random graphs, provides a simplified framework for developing theoretical tools and insights into landmark-based embedding algorithms. Particularly, Algorithm \ref{alg} achieves $(1\pm\varepsilon)$-factor approximations of shortest-path distances on random graphs w.h.p. even with reduced embedding dimensionality, complementing Bourgain's worst-case results \yrcite{Bourgain85}. By integrating GNNs into the embedding construction, we further improve its generalizability and transferability while reducing time and space complexity, as demonstrated by experiments on ER graphs and benchmark datasets. These signal the potential of machine learning-based landmark algorithms as a solution to graph representation learning for large, complex network data.

\noindent\textbf{Limitations and Future Work.} While our results improve upon existing landmark-based algorithms, several limitations remain. Our theoretical analysis focuses on ER graphs, a simplified model; extending it to more complex graphs (e.g., inhomogeneous random graphs, configuration models, planar graphs) is a key direction for future work. The approach also relies on GNNs generalizing from smaller to larger graphs; further studies are needed to assess robustness across diverse graph properties. Finally, additional improvements in memory and inference efficiency may be possible with advanced GNN architectures or alternative models.








\newpage
\appendix

\section{Proofs} \label{app:theory}

\subsection{Proof of Lemma \ref{lem:event-equivalence}}\label{app:pf3.5}

If $A_n$ occurs but $B_n$ does not, then $|\mathcal{C}_{\sss (2)}| \geq n^{\kappa_0 -\varepsilon}$, which occurs with probability tending to zero since $|\mathcal{C}_{\sss (2)}| = O(\log n)$ w.h.p. On the other hand, if $B_n$ occurs and $A_n$ does not, then $\smash{|\partial N_L(u_1)|\notin [n^{\kappa_0-\varepsilon}, n^{\kappa_0+\varepsilon}]}$ or $|\partial N_L(u_2)|\notin [n^{\kappa_0-\varepsilon}, n^{\kappa_0+\varepsilon}]$. 
  
To bound the probabilities of these events, consider a branching process with progeny distribution $\text{Poisson}(\lambda)$, and let $\mathcal{X}_l$ be the number of children at generation $l$.  
We first claim that, for any fixed node $u$ in $G$,  
\begin{equation}\label{eq:coupling}
\lim_{n\to\infty} \PR(|\partial N_L(u)| = \mathcal{X}_L) = 1 
\end{equation}  
for any $\kappa_0 \in (0, \frac{1}{2})$ and $L = \kappa_0 \log_\lambda n$. Indeed, this is a consequence of Lemma 3.13 from \newcite{Bordenave}.  

Next, classical theory of branching processes shows that, on the event of survival, the growth rate of a branching process is exponential. More precisely, Theorem 5.5 (iii) from \newcite{Tanny77}, together with Theorem 2 from \newcite{AN72}, yields  
\begin{align*}
\lim_{L\to\infty} \PR\big(L(1-\varepsilon) \leq \log_\lambda \mathcal{X}_L \leq L(1+\varepsilon), \, \mathcal{X}_L>0\big) = 1.
\end{align*}  

Since $\kappa_0 - \varepsilon < \kappa_0(1-\varepsilon)$ and $\kappa_0 + \varepsilon > \kappa_0(1+\varepsilon)$,  
\begin{equation}\label{eq:growth-of-BP}
\lim_{L\to\infty} \PR\big(n^{\kappa_0-\varepsilon} \leq \mathcal{X}_L \leq n^{\kappa_0+\varepsilon}, \, \mathcal{X}_L>0\big) = 1.
\end{equation}

Combining \eqref{eq:coupling} and \eqref{eq:growth-of-BP}, it follows that 
\begin{align*}
\PR(B_n\setminus A_n) \leq \PR(|\partial N_L(u_1)|\notin [n^{\kappa_0-\varepsilon}, n^{\kappa_0+\varepsilon}]) + \PR(|\partial N_L(u_2)|\notin [n^{\kappa_0-\varepsilon}, n^{\kappa_0+\varepsilon}])\to 0. 
\end{align*}

\subsection{Proof of Lemma \ref{lem:concentration-neighbohoods}}\label{app:pf3.4}

The proof is adapted from Section 2.6.4 in \newcite{RGCN2}. Since we need an exponential bound on the probability and $L$ grows with $n$, the proof does not follow from \newcite{RGCN2}.

Note that for any fixed node $u$ and any $k\geq 1$, $|\partial N_{k}(u)| \leq \sum_{x\notin N_{k-1} (u)}\sum_{y\in \partial N_{k-1} (u)} I_{xy}$, where $I_{xy}$ is the indicator random variable for the edge $\{x,y\}$ being present. Therefore, $\smash{\E(|\partial N_{k}(u)|) \leq \lambda E(|\partial N_{k-1}(u)|)}$. Proceeding inductively, we have $\E(|\partial N_{k}(u)|) \leq \lambda^k$ and consequently, 
\begin{align}
    \E(|N_k(u)|) \leq \frac{\lambda^{k+1}-1}{\lambda-1} = O( \lambda^k). \label{first}
\end{align}
Then with Markov's inequality, there exists $\delta>0$ for any $\gamma\in (\kappa_0+\kappa,1)$ such that 
$$\PR(| N_k(u_i)|\geq n^\gamma)\leq \frac{O(\lambda^{k})}{n^\gamma}\leq \frac{O\left(n^{\kappa_0+\kappa}\right)}{n^\gamma}\leq n^{-\delta}$$
for $i=1,2$ and $k\leq (\kappa_0+\kappa)\log_{\lambda} n$ with sufficiently large $n$. Then for each fixed $k\leq (\kappa_0+\kappa)\log_{\lambda} n$,
\begin{align}
\PR(| N_k(u_i) | \leq n^\gamma: i = 1,2) \geq 1- \sum_{i=1,2} \PR(| N_k(u_i)|\geq n^\gamma) \geq 1-2n^{-\delta}. \label{markov2}
\end{align}

Let $\delta_n=n^{-\beta}$ with $0<\beta<\frac{\kappa_0-\varepsilon'}{2}$. Also define $$\smash{\bar{\cE}_{n,k}=\{|\partial N_{L+k}(u_i)| \in [b'_m(1-\delta_n)^k(1-n^{\gamma-1})^k\lambda^k, b'_M(1+\delta_n)^k \lambda^k]:i=1,2\}}$$ with $b'_m=n^{-\varepsilon'}\lambda^L$ and $b'_M=n^{\varepsilon'}\lambda^L$ for some $0<\varepsilon'<\min\{\varepsilon,1-\kappa_0-\kappa\}$. Conditionally on $A_{b_m,b_M}$,
\begin{align*}
    \E(|\partial &N_{L+k}(u_i)|\mid  N_{L+k-1}(u_i))=\E\left(\sum_{\substack{x \notin N_{L+k-1}(u_i)}}\1_{\{\exists y\in \partial N_{L+k-1}(u_i):I_{xy}=1\}}\mid  N_{L+k-1}(u_i)\right)\\
    &=(n-|N_{L+k-1}(u_i)|)\PR(\exists y\in \partial N_{L+k-1}(u_i):I_{xy}=1\mid  N_{L+k-1}(u_i);\: x \notin N_{L+k-1}(u_i))\\
    &=(n-|N_{L+k-1}(u_i)|)\left(1-\left(1-\frac{\lambda}{n}\right)^{|\partial N_{L+k-1}(u_i)|}\right).
\end{align*}

Since $\frac{\lambda}{n}\in [0,1]$, 
$$1 - |\partial N_{L+k-1}(u_i)|\frac{\lambda}{n}\leq \left(1 - \frac{\lambda}{n}\right)^{|\partial N_{L+k-1}(u_i)|} \leq 1 - |\partial N_{L+k-1}(u_i)|\frac{\lambda}{n} + \frac{|\partial N_{L+k-1}(u_i)|^2}{2}\left(\frac{\lambda}{n}\right)^2.$$

Conditionally on $\bar{\cE}_{n,k-1}$,
\begin{align*}
    |\partial N_{L+k-1}&(u_i)|\frac{\lambda}{n}\leq  n^{\varepsilon'}(1+\delta_n)^{k-1}\frac{\lambda^{L+k}}{n}\leq n^{\varepsilon'-1}(1+\delta_n)^{k-1} n^{\kappa_0+\kappa}\to 0 \text{ as } n\to \infty.
\end{align*}
Since $\left(|\partial N_{L+k-1}(u_i)|\frac{\lambda}{n}\right)^2$ vanishes faster than $|\partial N_{L+k-1}(u_i)|\frac{\lambda}{n}$, we have w.h.p. that $$\left(1 - \frac{\lambda}{n}\right)^{|\partial N_{L+k-1}(u_i)|}=1 - |\partial N_{L+k-1}(u_i)|\frac{\lambda}{n}$$ and so $$\E(|\partial N_{L+k}(u_i)|\mid  N_{L+k-1}(u_i))=(n-|N_{L+k-1}(u_i)|)|\partial N_{L+k-1}(u_i)|\frac{\lambda}{n}.$$

Conditionally on $\cap_{l=0}^{k-1}\bar{\cE}_{n,l}$ and $A_{b_m,b_M}$, from \eqref{markov2} we have with probability at least $1-2n^{-\delta}$ that
\begin{align*}
    \E(|\partial N_{L+k}(u_i)|\mid  N_{L+k-1}(u_i))\in \left[b'_m(1-\delta_n)^{k-1}(1-n^{\gamma-1})^{k}\lambda^{k}, b'_M(1+\delta_n)^{k-1} \lambda^{k}\right]
\end{align*}
since $1-n^{\gamma-1}\leq 1-\frac{|N_{L+k-1}(u_i)|}{n}\leq 1$ for $i=1,2$ with sufficiently large $n$. Denote this event $R_k$. The fact that $\PR(A)\leq \PR(A\mid B)+\PR(B^c)$ implies
\begin{align*}
    \PR(\bar{\cE}_{n,k}^c &\mid \cap_{l=0}^{k-1}\bar{\cE}_{n,l}, A_{b_m,b_M}) \leq \PR(\bar{\cE}_{n,k}^c \mid R_k, \cap_{l=0}^{k-1}\bar{\cE}_{n,l}, A_{b_m,b_M}) + 2n^{-\delta}.
\end{align*}
Using union bound and Chernoff-Hoeffding bound \citep[Theorem 1.1]{dubhashi2009concentration},
\begin{align*}
    \PR&(\bar{\cE}_{n,k}^c \mid R_k, \cap_{l=0}^{k-1}\bar{\cE}_{n,l}, A_{b_m,b_M})\\
    &\leq \sum_{i=1,2}  \PR(||\partial N_{L+k}(u_i)|-\E(|\partial N_{L+k}(u_i)|)|\geq \delta_n \E(|\partial N_{L+k}(u_i)|) \mid R_{k}, \cap_{l=0}^{k-1}\bar{\cE}_{n,l}, A_{b_m,b_M})\\
    &\leq \sum_{i=1,2} 2\exp\left(-\frac{\delta_n^2}{3}n^{-\varepsilon'}(1-\delta_n)^{k-1}(1-n^{\gamma-1})^{k} \lambda^{L+k}\right)\\
    &\leq 4\exp\left(-\frac{n^{-2\beta}}{3}n^{-\varepsilon'}(1-\delta_n)^{k-1}(1-n^{\gamma-1})^{k} n^{\kappa_0}\right).    
\end{align*}

Since $(1-\delta_n)^{k-1},(1- n^{\gamma-1})^k\to 1$ as $n\to \infty$ and $2\beta<\kappa_0-\varepsilon'$, $4\exp\left(-\frac{n^{-2\beta}}{3}n^{-\varepsilon'}(1-\delta_n)^{k-1}(1-n^{\gamma-1})^{k} n^{\kappa_0}\right)$ vanishes faster than $2n^{-\delta}$. Then with sufficiently large $n$, $\smash{\PR(\bar{\cE}_{n,k} \mid  \cap_{l=0}^{k-1}\bar{\cE}_{n,l}, A_{b_m,b_M})\geq 1-3n^{-\delta}}$. Proceed inductively,
\begin{align*}
    \PR&(\cap_{l=0}^{k}\bar{\cE}_{n,l}\mid A_{b_m,b_M})=\PR(\bar{\cE}_{n,k} \mid  \cap_{l=0}^{k-1}\bar{\cE}_{n,l}, A_{b_m,b_M})\dots \PR(\bar{\cE}_{n,1} \mid  \bar{\cE}_{n,0}, A_{b_m,b_M})\PR(\bar{\cE}_{n,0} \mid A_{b_m,b_M})\\
    &\geq (1-3n^{-\delta})\dots (1-3n^{-\delta})\PR(\bar{\cE}_{n,0} \mid A_{b_m,b_M})= (1-3n^{-\delta})^{k}\PR(\bar{\cE}_{n,0} \mid A_{b_m,b_M}).
\end{align*}

Since $b'_m(1-\delta_n)^k(1-n^{\gamma-1})^k\geq b_{m}$ and $b'_M(1+\delta_n)^k\leq b_{M}$, $\bar{\cE}_{n,0}\subseteq A_{b_m,b_M}$ and $\bar{\cE}_{n,k}\subseteq \cE_{n,k}$ for all $k\geq 0$. Hence, $\PR(\bar{\cE}_{n,0} \mid A_{b_m,b_M})=1$ and so $$\PR(\cap_{l=0}^{k}\cE_{n,l}\mid A_{b_m,b_M})\geq \PR(\cap_{l=0}^{k}\bar{\cE}_{n,l}\mid A_{b_m,b_M})\geq (1-3n^{-\delta})^{k}\geq 1-3kn^{-\delta}.$$

\subsection{Proof of Proposition \ref{prop:union-intersection}}\label{app_prop}

Recall all the notation from Lemma \ref{lem:concentration-neighbohoods} and its proof. Then for any $k_1, k_2$ such that $\smash{L < k_1, k_2 \leq k= (\kappa_0+\kappa)\log_{\lambda} n}$,
\begin{align*}
    \E&(|\partial N_{k_1}(u_1)\cap \partial N_{k_2}(u_2)| \mid N_{k_1}(u_1),N_{k_2-1}(u_2))\\
    &=\E\left(\sum_{x \in \partial N_{k_1}(u_1) \setminus N_{k_2-1}(u_2)}\1_{\{\exists y\in \partial N_{k_2-1}(u_2):I_{xy}=1\}}\mid N_{k_1}(u_1),N_{k_2-1}(u_2)\right)\\
    &=\left(|\partial N_{k_1}(u_1)|-\sum_{j\leq k_2-1}|\partial N_{k_1}(u_1)\cap \partial N_{j}(u_2)|\right)\left(1-\left(1-\frac{\lambda}{n}\right)^{|\partial N_{k_2-1}(u_2)|}\right).
\end{align*}

Since $\frac{\lambda}{n}\in [0,1]$, 
$$1 - |\partial N_{k_2-1}(u_2)|\frac{\lambda}{n}\leq \left(1 - \frac{\lambda}{n}\right)^{|\partial N_{k_2-1}(u_2)|} \leq 1 - |\partial N_{k_2-1}(u_2)|\frac{\lambda}{n} + \frac{|\partial N_{k_2-1}(u_2)|^2}{2}\left(\frac{\lambda}{n}\right)^2.$$

Conditionally on $u_1, u_2$ being in the same connected component, Lemmas \ref{lem:event-equivalence} and \ref{lem:concentration-neighbohoods} imply that with probability at least $1-3n^{-\delta}(\lfloor k \rfloor - \lfloor L \rfloor)\geq 1-3n^{-\delta}(\kappa\log_{\lambda} n+1)$ for some $\delta>0$, $\cap_{l=0}^{k-L}\cE_{n,l}$ occurs. Then with $\varepsilon'\in(0,\min\{\varepsilon,1-\kappa_0-\kappa\})$ ($\varepsilon'$ to be chosen later),
\begin{align*}
    |\partial N_{k_2-1}(u_2)|\frac{\lambda}{n} \leq n^{\frac{\varepsilon'}{2}}\frac{\lambda^{k_2}}{n}\leq n^{\frac{\varepsilon'}{2}-1}n^{\kappa_0+\kappa}\to 0 \text{ as } n \to \infty.
\end{align*}

Since $|\partial N_{k_2-1}(u_2)|\frac{\lambda}{n}$ vanishes and $\left(|\partial N_{k_2-1}(u_2)|\frac{\lambda}{n}\right)^2$ vanishes faster, we have w.h.p. that $\left(1 - \frac{\lambda}{n}\right)^{|\partial N_{k_2-1}(u_2)|}=1 - |\partial N_{k_2-1}(u_2)|\frac{\lambda}{n}$, and so
\begin{align}
    \E &\big(|\partial N_{k_1}(u_1) \cap \partial N_{k_2}(u_2)| \:\big|\: N_{k_1}(u_1), N_{k_2-1}(u_2)\big)\notag \\
    &= \left(|\partial N_{k_1}(u_1)| - \sum_{j \leq k_2 - 1} |\partial N_{k_1}(u_1) \cap \partial N_{j}(u_2)|\right) |\partial N_{k_2 - 1}(u_2)| \frac{\lambda}{n} \label{equality}.
\end{align}
Conditionally on $\cap_{l=0}^{k-L}\cE_{n,l}$,
\begin{align}
    \E \big(|\partial N_{k_1}(u_1) &\cap \partial N_{k_2}(u_2)| \:\big|\: N_{k_1}(u_1), N_{k_2-1}(u_2)\big)\leq n^{\frac{\varepsilon'}{2}} \lambda^{k_1}n^{\frac{\varepsilon'}{2}}\frac{\lambda^{k_2}}{n}\label{upper}\\
    &\leq n^{\varepsilon'}\lambda^{k_1}\frac{n^{\kappa_0+\kappa}}{n}\leq \lambda^{k_1}\frac{n^{-\gamma}}{7(\lfloor k \rfloor-\lfloor L \rfloor)}\notag
\end{align}
for all $L< k_2\leq k$ with $0<\gamma<\min\{\kappa_0,1-\kappa_0-\kappa\}$ and sufficiently large $n$. Here we choose $\varepsilon'$ small enough so that $\frac{\varepsilon'}{2}<\gamma$, $\frac{k_1}{\log_{\lambda}n}-\frac{\varepsilon'}{2}>\kappa_0$, and $k_1+k_2>(1+\varepsilon')\log_\lambda n$.

Let $A$ be the event that there exists $L< j\leq k_2$ such that $\left|\partial N_{k_1}(u_1) \cap \partial N_{j}(u_2)\right| \geq \lambda^{k_1} \frac{n^{-\gamma}}{\lfloor k \rfloor-\lfloor L \rfloor}$. Let $B$ be the event that $\E(|\partial N_{k_1}(u_1) \cap \partial N_{j}(u_2)| \mid N_{k_1}(u_1),N_{k_2-1}(u_2))\leq \lambda^{k_1} \frac{n^{-\gamma}}{7(\lfloor k \rfloor-\lfloor L \rfloor)}$ for all $L < j\leq k_2$. The fact that $\PR(A)\leq \PR(A\mid B)+\PR(B^c)$ implies
\begin{align*}
    \PR(A\mid N_{k_1}(u_1),N_{k_2-1}(u_2)) \leq \PR(A\mid B, N_{k_1}(u_1),N_{k_2-1}(u_2))+3n^{-\delta}(\kappa\log_{\lambda} n+1).
\end{align*}

By Theorem 2.8 and Corollary 2.4 from \newcite{JLR00} with union bound, 
\begin{align*}
    \PR&(A\mid B, N_{k_1}(u_1),N_{k_2-1}(u_2))\leq (\lfloor k\rfloor-\lfloor L\rfloor)\exp\left( - \lambda^{k_1} \frac{n^{-\gamma}}{\lfloor k \rfloor-\lfloor L \rfloor} \right)\\
    &< (\kappa\log_{\lambda}n+1) \exp \left(-
    \frac{n^{\kappa_0-\gamma}}{\lfloor k \rfloor-\lfloor L \rfloor}\right)= (\kappa\log_{\lambda}n+1) \exp \left(-n^{\gamma'}\right)
\end{align*}
for some $\gamma'=\kappa_0-\gamma>0$. It follows that
\begin{align}
    \PR\Bigg(A^c\mid N_{k_1}(u_1), N_{k_2-1}(u_2)\Bigg)&\geq 1-(\kappa\log_{\lambda}n+1) \exp \left(-n^{\gamma'}\right)-3n^{-\delta}(\kappa\log_{\lambda} n+1)\notag\\
    &\geq 1-4n^{-\delta}(\kappa\log_{\lambda} n+1) \label{prop-all}
\end{align}
for sufficiently large $n$ since $(\kappa\log_{\lambda}n+1)\exp \left(-n^{\gamma'}\right)$ vanishes faster than $3n^{-\delta}(\kappa\log_{\lambda} n+1)$.

Let $\gamma''\in\left(\kappa_0,\frac{k_1}{\log_{\lambda}n}-\frac{\varepsilon'}{2}\right)$. Markov's inequality and Lemma \ref{first} imply that there exists $\delta'>0$ such that 
\begin{align}
\PR(| N_L(u_2)|\geq n^{\gamma''})\leq \frac{O(\lambda^{L})}{n^{\gamma''}}\leq \frac{O\left(n^{\kappa_0}\right)}{n^{\gamma''}}\leq n^{-\delta'}\label{markov3}
\end{align}
for sufficiently large $n$.

Combining \eqref{equality}, \eqref{upper}, \eqref{prop-all}, \eqref{markov3} with Lemmas \ref{lem:event-equivalence}, \ref{lem:concentration-neighbohoods}, we have w.h.p. that
\begin{align}
    \E &\big(|\partial N_{k_1}(u_1) \cap \partial N_{k_2}(u_2)| \:\big|\: N_{k_1}(u_1), N_{k_2-1}(u_2)\big)\notag\\
    &\geq \left(|\partial N_{k_1}(u_1)| - \sum_{L<j \leq k_2 - 1} |\partial N_{k_1}(u_1) \cap \partial N_{j}(u_2)|-|N_{L}(u_2)|\right) |\partial N_{k_2 - 1}(u_2)| \frac{\lambda}{n}\notag\\
    &\geq \left(n^{-\frac{\varepsilon'}{2}}\lambda^{k_1} - (\lfloor k_2\rfloor -1-\lfloor L \rfloor) \lambda^{k_1} \frac{n^{-\gamma}}{\lfloor k \rfloor-\lfloor L \rfloor}-n^{\gamma''}\right) \frac{n^{-\frac{\varepsilon'}{2}}\lambda^{k_2}}{n}> \frac{n^{-\varepsilon'}\lambda^{k_1+k_2}}{2n}\notag
\end{align}
and 
\begin{align}
    \E \big(|\partial N_{k_1}&(u_1) \cap \partial N_{k_2}(u_2)| \:\big|\: N_{k_1}(u_1), N_{k_2-1}(u_2)\big)
    \leq n^{\varepsilon'}\frac{\lambda^{k_1+k_2}}{n},\notag
\end{align}
where the last "$\geq$" holds since $\lambda^{k_1} n^{-\gamma}$ and $n^{\gamma''}$ grow strictly slower than $n^{-\frac{\varepsilon'}{2}}\lambda^{k_1}$ as $\frac{\varepsilon'}{2}<\gamma$ and $\gamma''<\frac{k_1}{\log_{\lambda}n}-\frac{\varepsilon'}{2}$. Therefore, $\E(|\partial N_{k_1}(u_1) \cap \partial N_{k_2}(u_2)|) \in \left[\frac{n^{-\varepsilon'}\lambda^{k_1+k_2}}{2n}, \frac{n^{\varepsilon'}\lambda^{k_1+k_2}}{n}\right]$ and we denote this event $S$.

Let $R$ denote the event that $|\partial N_{k_1}(u_1) \cap \partial N_{k_2}(u_2)|\notin \left[\frac{(1-\varepsilon)n^{-\varepsilon'}\lambda^{k_1+k_2}}{2n}, \frac{(1+\varepsilon)n^{\varepsilon'}\lambda^{k_1+k_2}}{n}\right]$. Using Chernoff-Hoeffding bound \citep[Theorem 1.1]{dubhashi2009concentration},
\begin{align*}
    \PR(R) \leq \PR(R\mid S) + \PR(S^c) \leq 2\exp\left(-\frac{\varepsilon^2}{3}\frac{n^{-\varepsilon'}\lambda^{k_1+k_2}}{2n}\right)+\PR(S^c).
\end{align*}
Since $k_1+k_2>(1+\varepsilon')\log_\lambda n$ and $S$ occurs w.h.p., $\PR\left(R^c\right)$ converges to 1. Then w.h.p., $$|\partial N_{k_1}(u_1) \cap \partial N_{k_2}(u_2)| \in \left[\frac{(1-\varepsilon)n^{-\varepsilon'}\lambda^{k_1+k_2}}{2n}, \frac{(1+\varepsilon)n^{\varepsilon'}\lambda^{k_1+k_2}}{n}\right]\subseteq \left[\frac{n^{-\varepsilon}\lambda^{k_1+k_2}}{2n}, \frac{n^{\varepsilon}\lambda^{k_1+k_2}}{n}\right].$$

\subsection{Proof of Theorem \ref{thm:main_lb}}\label{app_lb}

Let $k_1 = \varepsilon' d(u_1,u_2)$ and $k_2 = (1- \varepsilon+\varepsilon') d(u_1,u_2)$, where $\smash{\varepsilon'=\min\left\{\frac{\varepsilon}{2},\varepsilon -\theta\right\}-\varepsilon''\in \left(0,\min\left\{\frac{\varepsilon}{2},\varepsilon -\theta\right\}\right)}$ ($\varepsilon''\in \left(0,\min\left\{\frac{\varepsilon}{2},\varepsilon -\theta\right\}\right)$ to be chosen later). Since $\varepsilon'<\frac{\varepsilon}{2}$, $k_1+k_2<d(u_1,u_2)$, and so $N_{k_1}(u_1) \cap N_{k_2}(u_2) = \varnothing$. Conditionally on $u_1,u_2$ being in the same connected component, Theorem 2.36 from \newcite{RGCN2} implies that $d(u_1,u_2)/ \log_{\lambda}  n \xrightarrow{\sss \PR} 1$. In other words, $(1-\epsilon)\log_{\lambda}  n\leq d(u_1,u_2)\leq (1+\epsilon)\log_{\lambda}  n $ w.h.p. for any fixed $\epsilon>0$. With $\epsilon$ small enough so that $\varepsilon'(1+\epsilon)<1$, $k_1\leq \varepsilon'(1+\epsilon)\log_{\lambda}  n < \log_{\lambda} n$ w.h.p., allowing us to apply Lemmas \ref{lem:event-equivalence} and \ref{lem:concentration-neighbohoods} on $|\partial N_{k_1}(u_1)|$. 

Let $S_{ij}$ be the landmark set of size $M^i$ sampled in the $j$-th round and $Z_{ij}$ denote the event that $\smash{S_{ij}\cap N_{k_1}(u_1) \neq \varnothing}$ but $S_{ij}\cap N_{k_2}(u_2) = \varnothing$. If $Z_{ij}$ happens for some $i\leq r$ and $j\leq R$, then $d(u_1,S_{ij}) \leq k_1$ and $d(u_2,S_{ij}) \geq k_2$, and consequently, $\smash{\underline{d} (u_1,u_2) \geq k_2-k_1= (1-\varepsilon) d(u_1,u_2)}$. Thus, denoting $Z = \cup_{i\leq r, j\leq R} Z_{ij}$, it suffices to prove that $\smash{\PR(Z \mid u_1 \leftrightarrow u_2) \xrightarrow{\sss \PR} 1}$. Since $\smash{\PR(u_1 \leftrightarrow u_2 \text{ but } u_1,u_2\notin \mathcal{C}_{\sss (1)}) = \frac{1}{n^2}\sum_{i\geq 2} |\mathcal{C}_{\sss (i)}|^2 \leq \frac{|\mathcal{C}_{\sss (2)}|}{n} \xrightarrow{\sss \PR} 0 }$, it suffices to show that $\PR(Z \mid u_1, u_2\in \mathcal{C}_{\sss (1)} ) \xrightarrow{\sss \PR} 1$ (or equivalently $\PR(Z^c \mid u_1, u_2\in \mathcal{C}_{\sss (1)} )\xrightarrow{\sss \PR} 0$). The fact that $\PR(A^c \cap B) = \PR(B) - \PR(A\cap B)$ implies, for each $(i,j)$, that 
\begin{align*}
  \PR(Z_{ij}\mid u_1, u_2\in \mathcal{C}_{\sss (1)}) &= \PR(S_{ij}\cap N_{k_1}(u_1)\neq \varnothing,S_{ij}\cap N_{k_2}(u_2)= \varnothing \mid u_1,u_2 \in \mathcal{C}_{\sss (1)}) \\
  &= \bigg(1-\frac{|N_{k_2}(u_2)|}{n}\bigg)^{M^i} - \bigg(1-\frac{|N_{k_1}(u_1)| + |N_{k_2}(u_2)|}{n}\bigg)^{M^i}. 
\end{align*}
By independence of $Z_{ij}$'s,
\begin{align*}
  &\PR(Z^c\mid u_1, u_2\in \mathcal{C}_{\sss (1)})\\
  &=\bigg(\prod_{i=0}^{r}\bigg(1-\bigg(1-\frac{|N_{k_2}(u_2)|}{n}\bigg)^{M^i} + \bigg(1-\frac{|N_{k_1}(u_1)| + |N_{k_2}(u_2)|}{n}\bigg)^{M^i}\bigg)\bigg)^R\notag\\
  &\leq \exp\bigg(-R\sum_{i=0}^r \bigg(\bigg(1-\frac{|N_{k_2}(u_2)|}{n}\bigg)^{M^i} - \bigg(1-\frac{|N_{k_1}(u_1)| + |N_{k_2}(u_2)|}{n}\bigg)^{M^i}\bigg)\bigg)\notag\\
  & = \exp\bigg(-R \sum_{i=0}^r \frac{|N_{k_1}(u_1)|}{n} \sum_{j=0}^{M^i-1} \bigg(1-\frac{|N_{k_2}(u_2)|}{n}\bigg)^{M^i - 1 - j} \bigg(1-\frac{|N_{k_1}(u_1)| + |N_{k_2}(u_2)|}{n}\bigg)^{j} \bigg)\notag\\
  & \leq \exp\bigg(-R \frac{| N_{k_1}(u_1)|}{n} \sum_{i=0}^r M^i \bigg(1-\frac{|N_{k_1}(u_1)| + |N_{k_2}(u_2)|}{n}\bigg)^{M^i-1} \bigg)\notag\\
  &< \exp\bigg(-R \frac{|\partial N_{k_1}(u_1)|}{n} M^{r} \bigg(1-\frac{|N_{k_1}(u_1)| + |N_{k_2}(u_2)|}{n}\bigg)^{M^r} \bigg)\notag
\end{align*}
where the first "$\leq$" uses $1-x\leq \exp(-x)$ and "$<$" uses $\sum_{i=0}^r M^i=\frac{M^{r+1}-1}{M-1}>\frac{M^{r+1}-M^{r}}{M-1}=M^r$.

Recall that $(1-\epsilon)\log_{\lambda} n\leq d(u_1,u_2)\leq (1+\epsilon)\log_{\lambda} n $ w.h.p. for any fixed $\epsilon>0$. Choosing $\epsilon$ small enough so that $(1-\varepsilon+\varepsilon')(1+\epsilon)<1-\theta$, we have that $k_2\leq (1-\varepsilon+\varepsilon')(1+\epsilon)\log_{\lambda} n<\left(1-\theta\right)\log_{\lambda} n$ w.h.p., and so there exists $\gamma\in \left(0,1-\theta\right)$ such that $k_1<k_2<\gamma\log_{\lambda} n$. By Markov's inequality and \eqref{first}, there exists $\delta>0$ such that 
$\PR(| N_{k_i}(u_i)|\geq n^\gamma)\leq \frac{O(\lambda^{k_i})}{n^\gamma}\leq n^{-\delta}$
for $i=1,2$ with sufficiently large $n$. Therefore,
\begin{align*}
\PR(| N_{k_i}(u_i) | &\leq n^\gamma: i = 1,2) \geq 1- \sum_{i=1,2} \PR(| N_{k_i}(u_i)|\geq n^\gamma) \geq 1-2n^{-\delta},
\end{align*}
and so $|N_{k_i}(u_i)|\leq n^\gamma$ for $i=1,2$ w.h.p. 

By Lemmas \ref{lem:event-equivalence} and \ref{lem:concentration-neighbohoods}, $|\partial N_{k_1}(u_1)|\geq n^{-\varepsilon'''}\lambda^{k_1}\geq n^{-\varepsilon'''}n^{\varepsilon'(1-\epsilon)}$ w.h.p. for any $\varepsilon'''>0$, and so
\begin{align*}
  \PR&(Z^c\mid u_1, u_2\in \mathcal{C}_{\sss (1)}) \\
  &< \exp\bigg(-R \frac{n^{-\varepsilon'''}n^{\left(\min\left\{\frac{\varepsilon}{2},\varepsilon -\theta\right\}-\varepsilon''\right)(1-\epsilon)}}{n} M^{\frac{\theta}{\log M} \log n-1} \bigg(1-\frac{2n^\gamma}{n}\bigg)^{M^{\frac{\theta}{\log M} \log n}} \bigg)\notag\\
  &= \exp\bigg(-R \frac{n^{-\varepsilon'''}n^{\min\left\{\frac{\varepsilon}{2},\varepsilon -\theta\right\}-\epsilon\min\left\{\frac{\varepsilon}{2},\varepsilon -\theta\right\}-\varepsilon''+\varepsilon''\epsilon}}{nM} n^{\theta} \bigg(1-\frac{2n^\gamma}{n}\bigg)^{n^{\theta}} \bigg).
\end{align*}
Since $\gamma<1-\theta$, $\bigg(1-\frac{2n^{\gamma}}{n}\bigg)^{n^{\theta}} \geq 1-\frac{2n^{\gamma+\theta}}{n} \to 1$ as $n\to \infty$. Since $\varepsilon'',\varepsilon''',\epsilon$ can be chosen small enough so that $-\varepsilon'''-\epsilon\min\left\{\frac{\varepsilon}{2},\varepsilon -\theta\right\}-\varepsilon''+\varepsilon''\epsilon<\varsigma$ for any $\varsigma>0$, $R=\Omega\left(M n^{1-\theta-\min\left\{\frac{\varepsilon}{2},\varepsilon -\theta\right\}+\varsigma}\right)$ is sufficient for the final bound to tend to 0. Since $\theta\in (0,\varepsilon)$, $R$ can be further simplified to $\Omega\left(M n^{1-\frac{\varepsilon}{2}-\min\left\{\frac{\varepsilon}{2},\theta\right\}+\varsigma}\right)$.

\subsection{Proof of Theorem \ref{thm:main_ub}}\label{app_ub}

Let $k =\varepsilon' d(u_1,u_2)$ with $\varepsilon'=\frac{1+\varepsilon}{2}-\varepsilon''\in\left(0,\frac{1+\varepsilon}{2}\right)$ ($\varepsilon''\in \left(0,\frac{1+\varepsilon}{2}\right)$ to be chosen later). Conditionally on $u_1,u_2$ being in the same connected component, Theorem 2.36 from \newcite{RGCN2} implies that $d(u_1,u_2)/ \log_{\lambda} n \xrightarrow{\sss \PR} 1$. In other words, $\smash{(1-\epsilon)\log_{\lambda}  n\leq d(u_1,u_2)\leq (1+\epsilon)\log_{\lambda} n }$ w.h.p. for any fixed $\epsilon>0$. With $\varepsilon'',\epsilon$ small enough so that $\varepsilon'(1+\epsilon)<1$ and $2\left(\frac{1+\varepsilon}{2}-\varepsilon''\right)(1-\epsilon)>1$, $k\leq \varepsilon'(1+\epsilon)\log_{\lambda}n<\log_{\lambda}n$ and $k+k\geq 2\varepsilon'(1-\epsilon)\log_{\lambda}n>\log_{\lambda}n$ w.h.p. This allows us to apply Lemmas \ref{lem:event-equivalence} and \ref{lem:concentration-neighbohoods} and Proposition \ref{prop:union-intersection}.

Let $S_{ij}$ be the landmark set of size $M^i$ sampled in the $j$-th round and $Z_{ij}$ be the event that $S_{ij}$ contains at least one landmark node in $N_k(u_1)\cap N_k(u_2)$ and none in $\smash{(N_k(u_1)\cup N_k(u_2))\setminus (N_k(u_1)\cap N_k(u_2))}$. If $Z_{ij}$ happens for some $i\leq r$ and $j\leq R$, the landmarks in the intersection will be the common landmarks for calculating $\bar{d}(u_1,u_2)$, and so $\bar{d}(u_1,u_2) \leq 2k\leq (1+\varepsilon) d(u_1,u_2)$. Thus, denoting $\smash{Z = \cup_{i\leq r, j\leq R} Z_{ij}}$, it suffices to prove that $\PR(Z \mid u_1 \leftrightarrow u_2) \xrightarrow{\sss \PR} 1$. Since $\smash{\PR(u_1 \leftrightarrow u_2 \text{ but } u_1,u_2\notin \mathcal{C}_{\sss (1)}) = \frac{1}{n^2}\sum_{i\geq 2} |\mathcal{C}_{\sss (i)}|^2 \leq \frac{|\mathcal{C}_{\sss (2)}|}{n} \xrightarrow{\sss \PR} 0}$, it suffices to show that $\smash{\PR(Z \mid u_1, u_2\in \mathcal{C}_{\sss (1)} ) \xrightarrow{\sss \PR} 1}$ (or equivalently $\PR(Z^c \mid u_1, u_2\in \mathcal{C}_{\sss (1)} )\xrightarrow{\sss \PR} 0$). Note that for each (i,j),
\begin{align*}
    \PR&(Z_{ij} \mid u_1, u_2\in \mathcal{C}_{\sss (1)})\\
    &=\frac{|N_k(u_1)\cap N_k(u_2)|}{n}\left(\frac{|N_k(u_1)\cap N_k(u_2)|}{n}+1-\frac{|N_k(u_1)\cup N_k(u_2)|}{n}\right)^{M^i-1}.
\end{align*}
By independence of $Z_{ij}$'s,
{\small \begin{align*}
    &\PR(Z^c \mid u_1, u_2\in \mathcal{C}_{\sss (1)})\\ &=\left(\prod_{i=0}^r\left(1-\frac{|N_k(u_1)\cap N_k(u_2)|}{n}\left(\frac{|N_k(u_1)\cap N_k(u_2)|}{n}+1-\frac{|N_k(u_1)\cup N_k(u_2)|}{n}\right)^{M^i-1}\right)\right)^R\\
    &\leq \exp\left(-R\sum_{i=0}^{r}\frac{|\partial N_k(u_1)\cap \partial N_k(u_2)|}{n}\left(\frac{|\partial N_k(u_1)\cap \partial N_k(u_2)|}{n}+1-\frac{|N_k(u_1)|+|N_k(u_2)|}{n}\right)^{M^i-1}\right).
\end{align*}}

Choosing $L\in(0,\min\{k,\gamma\log_{\lambda} n\})$ for some $\gamma\in \left(0,1-\theta\right)$, we obtain from Markov's inequality and Lemma \ref{first} that $\PR(| N_L(u_i)|\geq n^\gamma)\leq \frac{O(\lambda^{L})}{n^\gamma}\leq n^{-\delta}$ for $i=1,2$ with some $\delta>0$ and sufficiently large $n$. Therefore,
\begin{align*}
\PR(| N_L(u_i) | &\leq n^\gamma: i = 1,2) \geq 1- \sum_{i=1,2} \PR(| N_L(u_i)|\geq n^\gamma) \geq 1-2n^{-\delta},
\end{align*}
and so $|N_{L}(u_i)|\leq n^\gamma$ for $i=1,2$ w.h.p. Then by Lemmas \ref{lem:event-equivalence} and \ref{lem:concentration-neighbohoods}, 
\begin{align*}
    |N_{k}(u_i)|&=|N_{L}(u_i)|+\sum_{l=L+1}^{k}|\partial N_{l}(u_i)|\leq n^\gamma+\sum_{l=L+1}^{k}n^{\varepsilon'''}\lambda^l < n^\gamma+(\log_{\lambda}n) n^{\varepsilon'''}\lambda^k
\end{align*}
for $i=1,2$ w.h.p. with $0<\varepsilon'''<1-\kappa_0-\kappa$. Combining these with Proposition \ref{prop:union-intersection}, we have w.h.p. that
\begin{align*}
    \PR&(Z^c \mid u_1, u_2\in \mathcal{C}_{\sss (1)})\\
    &\leq \exp\left(-R\frac{n^{-\varepsilon'''}\lambda^{2k}}{2n^2}\sum_{i=0}^{r}\left(\frac{n^{-\varepsilon'''}\lambda^{2k}}{2n^2}+1-\frac{2n^\gamma}{n}-\frac{2(\log_{\lambda}n) n^{\varepsilon'''}\lambda^k}{n}\right)^{M^i-1}\right).
\end{align*}

Since $\gamma<1$ and $0<\lambda^{k}\leq n^{\kappa_0+\kappa}<n^{1-\varepsilon'''}<n$, $\frac{2n^\gamma}{n}+\frac{2(\log_{\lambda}n) n^{\varepsilon'''}\lambda^k}{n}-\frac{n^{-\varepsilon'''}\lambda^{2k}}{2n^2}\in (0,1)$ when $n$ is large, and so
\begin{align*}
    \PR&(Z^c \mid u_1, u_2\in \mathcal{C}_{\sss (1)})\\
    &\leq \exp\left(-R\frac{n^{-\varepsilon'''}\lambda^{2k}}{2n^2}\sum_{i=0}^{r}\left(1-\left(\frac{2n^\gamma}{n}+\frac{2(\log_{\lambda}n) n^{\varepsilon'''}\lambda^k}{n}-\frac{n^{-\varepsilon'''}\lambda^{2k}}{2n^2}\right)(M^i-1)\right)\right)\\
    &<\exp\left(-R\frac{n^{-\varepsilon'''}\lambda^{2k}}{2n^2}\left(r -\left(\frac{2n^\gamma}{n}+\frac{2(\log_{\lambda}n) n^{\varepsilon'''}\lambda^k}{n}-\frac{n^{-\varepsilon'''}\lambda^{2k}}{2n^2}\right)\frac{n^\theta M}{M-1}\right)\right)
\end{align*}
since $\sum_{i=0}^r(M^i-1)< \sum_{i=0}^r M^i=\frac{M^{r+1}-1}{M-1}\leq \frac{n^\theta M}{M-1}$. Since $\gamma<1-\theta$, $0<\frac{2n^{\gamma+\theta}}{n}\to 0$ as $n\to \infty$. Since we can choose $\varepsilon'',\varepsilon''',\epsilon$ small enough so that $\varepsilon'''+\left(\frac{1+\varepsilon}{2}-\varepsilon''\right)(1+\epsilon)<1-\theta$, we then obtain $0<n^{-\varepsilon'''}\frac{\lambda^{2k}}{n^2}n^{\theta}< n^{\varepsilon'''}\frac{\lambda^{k}}{n}n^{\theta}\leq \frac{n^{\varepsilon'''+\left(\frac{1+\varepsilon}{2}-\varepsilon''\right)(1+\epsilon)+\theta}}{n}<1$ for sufficiently large $n$. Then w.h.p.,
\begin{align*}
    \PR(Z^c \mid u_1, u_2\in \mathcal{C}_{\sss (1)})&< \exp\left(-R\frac{n^{-\varepsilon'''}n^{2\left(\frac{1+\varepsilon}{2}-\varepsilon''\right)(1-\epsilon)}}{2n^2}\left(\frac{\theta}{\log M}\log n-1-\left(1+1-0\right)\right)\right)\\
    &< \exp\left(-R\frac{n^{-\varepsilon'''}n^{1+\varepsilon+2\left(\varepsilon''\epsilon-\varepsilon''-\epsilon\frac{1+\varepsilon}{2}\right)}}{2n^2}\frac{\theta}{2\log M}\log n\right).\label{prop-ub}\\
\end{align*}
Since $\varepsilon'',\varepsilon''',\epsilon$ can be chosen small enough so that $-\varepsilon'''+2\left(\varepsilon''\epsilon-\varepsilon''-\epsilon\frac{1+\varepsilon}{2}\right)<\varsigma$ for any $\varsigma>0$, $R=\Omega\left(\frac{\log M}{\theta\log n}n^{1-\varepsilon+\varsigma}\right)$ is sufficient for the final bound to tend to 0.

\section{Experimental Setup} \label{app:exp_details}

In our experiments, we train GNNs to approximate the landmark distances in sparse, undirected, unweighted random graphs. We consider four standard GNN architectures (GCN \citep{kipf17-classifgcnn}, GraphSAGE \citep{NIPS2017_5dd9db5e}, GAT \citep{Velickovic18-GraphAttentionNetworks}, and GIN \citep{xu2018how}) with sum aggregation, dropout, and ReLU activations. For each architecture, we test nine models with $\lfloor \sqrt{n} \rfloor$ nodes in the first and last layers and hidden layers varying in depth and width:
\begin{itemize}
  \item Depth-6: 128-64-32-16, 64-32-16-8, 32-16-8-4
  \item Depth-5: 128-64-32, 64-32-16, 32-16-8
  \item Depth-4: 128-64, 64-32, 32-16
\end{itemize}

The training data for the GNNs are graphs generated by $\ER(\lambda/n)$ with $1 < \lambda \ll n$, which ensures sparsity and the existence of a giant component w.h.p. In particular, we consider $\lambda \in \{3,4,5,6\}$ and $n \in \{25,50,100,200,400,800,1600,3200\}$. Each graph is treated as a batch of nodes with a 200-50-50 train-validation-test split to generate random input signals $\bbX \in \reals^{n \times r}$, where each column one-hot encodes a landmark node. The outputs $\bbY \in \reals^{n \times r}$ have the same dimensions as the inputs and represent shortest path distances between nodes $u$ and landmarks $s$, i.e., $[\bbY]_{us} = d(u,s)$. Training runs for 1000 epochs with early stopping (100 epochs), MSE loss, Adam optimizer (lr=0.01, weight decay=0.0001), and a cyclic-cosine learning rate schedule (0.001–0.1 for 10 cycles, with default cosine annealing for up to 20 iterations).

All experiments use PyTorch Geometric \citep{fey2019fast} on a Lambda Vector 1 machine (AMD Ryzen Threadripper PRO 5955WX CPU, 16 cores, 128 GB RAM, 2× NVIDIA RTX 4090 GPUs, no parallel training). Code is available at \url{https://github.com/ruiz-lab/shortest-path}.

\section{More Experimental Results} \label{app:more_exps}

We provide additional results with more detailed explanations offering deeper insights into GNNs and their use for generating landmark embeddings in shortest path approximations. The experiments are divided into three categories: learning the predictive power of GNNs, comparing the performance of the GNN-augmented approach with the vanilla landmark-based algorithm, and evaluating the transferability of both methods to larger random graphs and real-world benchmarks.

\subsection{Experiment 1: Learning the GNNs}\label{exp1app}

\begin{figure*}[h]
\centering
\includegraphics[width=0.4\textwidth]{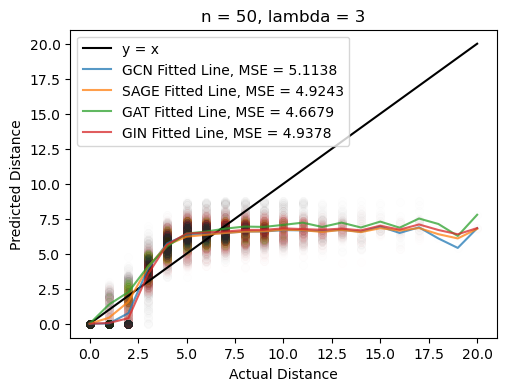}
\includegraphics[width=0.4\textwidth]{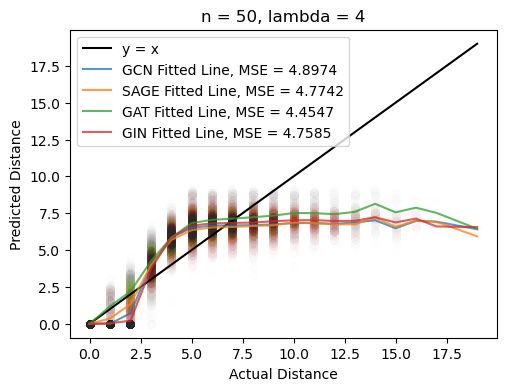}\\
\:\includegraphics[width=0.4\textwidth]{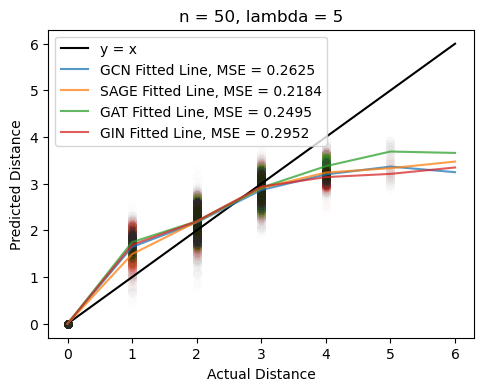}\:
\includegraphics[width=0.4\textwidth]{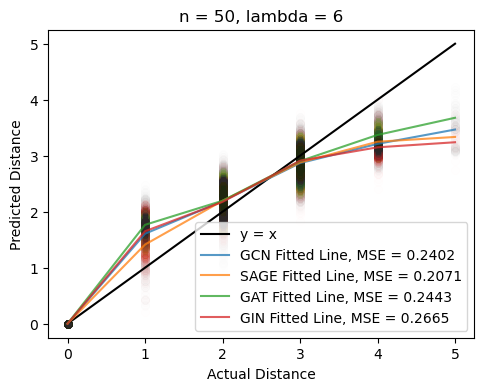}
\caption{End-to-end shortest path distance predictions from $\lfloor \sqrt{n}\rfloor\text{-64-32-16-}\lfloor\sqrt{n}\rfloor$ GNNs trained on graphs generated by $\ER(\lambda/n)$. The evaluation data consists of graphs from the same model.}
\label{fig:exp1app}
\end{figure*}

In the first experiment, we evaluate the ability of trained GNNs to compute end-to-end shortest paths. We consider $n=50$ and set the GNN depth to be larger than $\lceil \log_\lambda n \rceil$. Figure~\ref{fig:exp1app} plots the actual shortest path distances versus those predicted by our selected GNN architectures. Predictions for distances beyond the GNN depth saturate, indicating that GNNs cannot capture longer distances even with depth exceeding the expected path length. As expected, GNNs are not suitable for computing end-to-end shortest path distances, especially on sparser graphs with $\lambda \in \{3,4\}$, which tend to exhibit longer paths.

\subsection{Experiment 2: Comparing BFS-Based and GNN-Based Landmark Embeddings}\label{exp2app}

\begin{figure*}[h]
\centering
\includegraphics[width=1\textwidth]{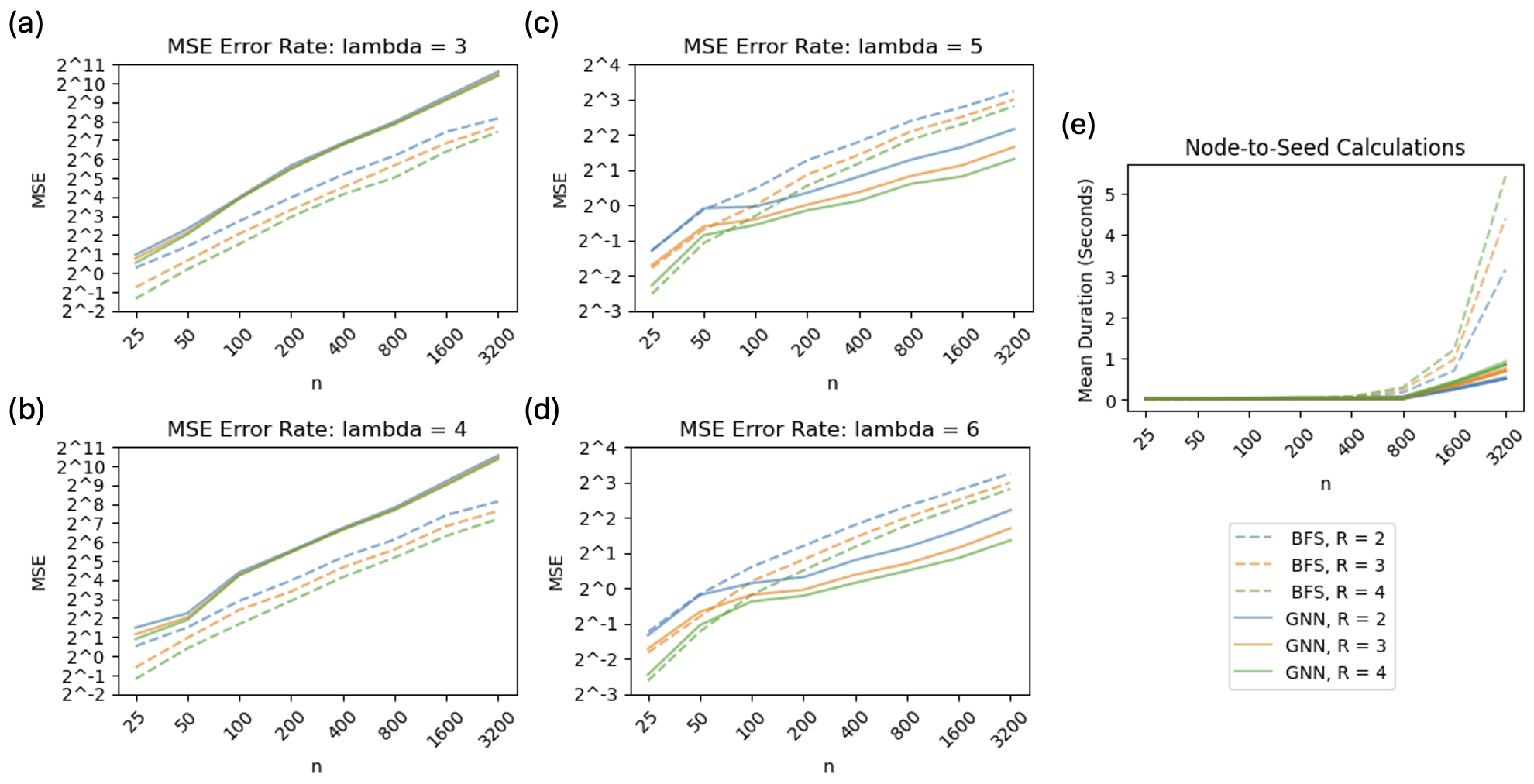}
\caption{Error rates of BFS-based and GNN-based lower bounds on graphs generated by $\ER(\lambda/n)$, with the GNNs trained on graphs from the same model.}
\label{fig:exp2app}
\end{figure*}

In this experiment, we compare the lower bounds (LBs) resulting from BFS-based and GNN-based landmark embeddings against the actual shortest path distances. \textit{Only LBs are compared to ensure a fair evaluation, as computing the upper bounds (UBs) requires storing additional information—namely, the indices of the closest landmarks from the landmark sets to each node. Moreover, unlike in LB computations, the saturation effect inherent in GNNs cannot be mitigated in UB computations, making the UB an unreliable metric for shortest path approximation when calculated upon GNN-based landmark distances.}

To construct the landmark embeddings, we sample $r+1$ landmark sets $S_0, S_1, \dots, S_r$ of cardinalities $2^0, 2^1, \dots, 2^r$ with $r = \lfloor \log n \rfloor$ for $R$ repetitions. In Figure~\ref{fig:exp2app}(a-d), GNN-based lower bounds underperform the vanilla lower bounds for smaller $\lambda \in \{3,4\}$, but yield substantial improvements for larger $\lambda \in \{5,6\}$ across all three tested values of $R$. Although both $\lambda$ values are in the supercritical regime ($\lambda > 1$), several factors explain this difference. As shown in Figure~\ref{fig:exp1app}, the GNN learns poorer landmark embeddings for $\lambda\in \{3,4\}$, even on small 50-node graphs. Additionally, for large $n$, graphs are almost surely connected when $\lambda\in \{5,6\}$ but not when $\lambda\in \{3,4\}$. Finally, Figure~\ref{fig:exp2app}(e) illustrates that GNN-based embeddings can be generated faster than BFS-based embeddings, particularly on large graphs as exact local embedding computations via BFS scale poorly with graph size.

\subsection{Experiment 3: Transferability}\label{exp3app}

In our last experiment, we investigate whether GNNs trained on small graphs can be transferred to compute landmark embeddings on larger networks for downstream shortest path approximation via LBs. This is motivated by \newcite{ruiz20-transf} and \newcite{ruiz2021transferability}, which show that GNNs are transferable as their outputs converge on convergent graph sequences. This, in turn, allows models trained on smaller graphs to generalize to similar larger graphs.

\begin{figure*}[h]
\centering
\includegraphics[width=1\textwidth]{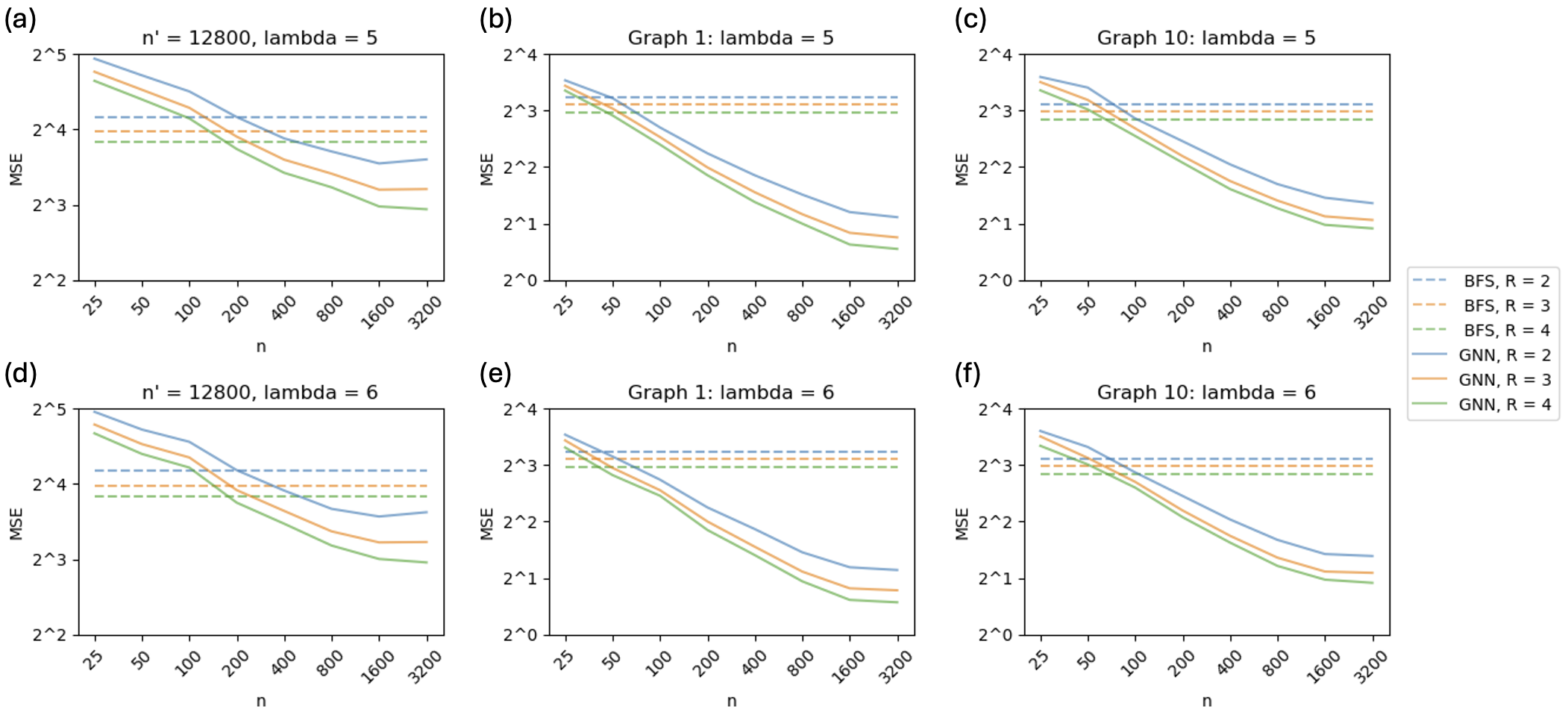}
\caption{Error rates of BFS-based and GNN-based lower bounds on (a,d) test ER graphs generated by $\text{ER}_{n'}(\lambda/n')$, (b,e) Arxiv COND-MAT collaboration network with 21,364 nodes, and (c,f) GEMSEC company network with 14,113 nodes, with the GNNs trained on graphs from $\ER(\lambda/n)$.}
\label{fig:exp3app}
\end{figure*}

Here, we focus on $\lambda \in \{5,6\}$ and train a sequence of eight GNNs on ER graphs ranging from $n=25$ to $n=3200$ nodes. These GNNs are then used to generate local node embeddings on graphs from the ER model with the same $\lambda$ and $n' = 12800$ nodes. Figure~\ref{fig:exp3app}(a,d) shows the MSE for each instance as the training graph size increases, with flat dashed lines indicating the MSE of BFS-based LBs on the $n'$-node graph. We observe a steady decrease in MSE as $n$ grows, with GNN-based embeddings matching BFS-based performance when trained on graphs of $n=100$, which is 128 times smaller than the target graph.

\begin{figure*}
\centering
\includegraphics[width=0.3\textwidth]{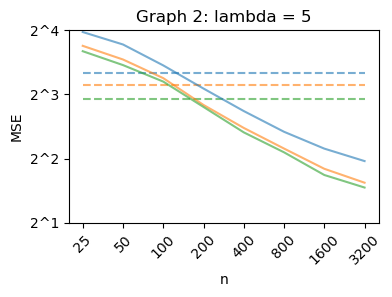}
\includegraphics[width=0.3\textwidth]{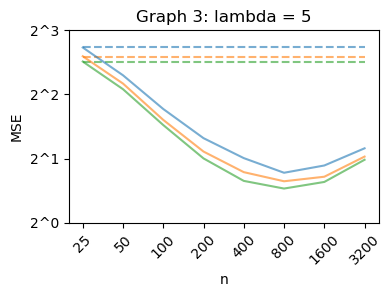}
\includegraphics[width=0.3\textwidth]{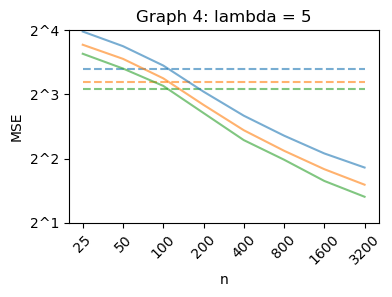}\\
\includegraphics[width=0.3\textwidth]{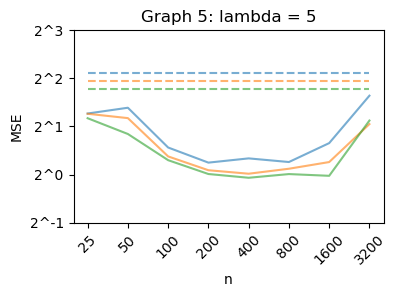}
\includegraphics[width=0.3\textwidth]{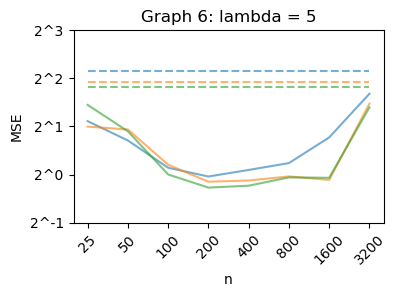}
\includegraphics[width=0.3\textwidth]{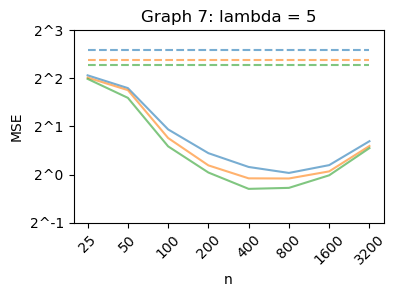}\\
\includegraphics[width=0.3\textwidth]{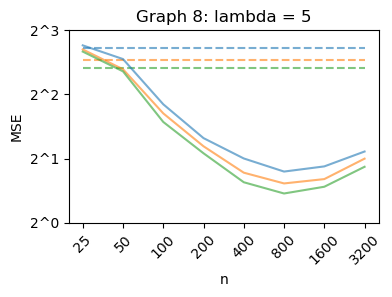}
\includegraphics[width=0.3\textwidth]{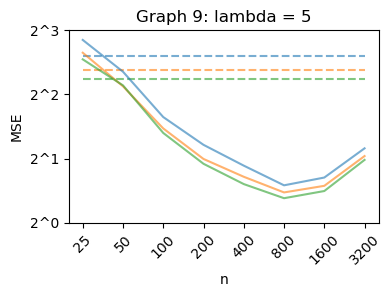}
\includegraphics[width=0.3\textwidth]{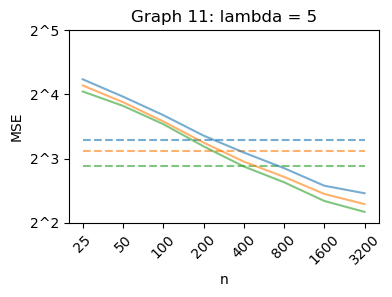}\\
\includegraphics[width=0.3\textwidth]{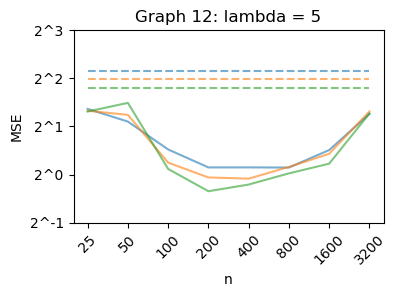}
\includegraphics[width=0.3\textwidth]{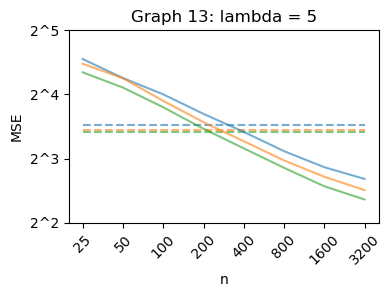}
\includegraphics[width=0.3\textwidth]{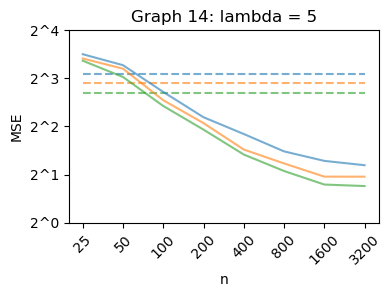}
\caption{Additional transferability results on real networks, with the GNNs trained on graphs from $\ER(\lambda/n)$. Legend is the same as in Figure \ref{fig:exp3app}.}
\label{fig:exp3app2}
\end{figure*}

When examining the transferability of the same set of GNNs on sixteen real-world networks listed in Table~\ref{tab}, we again observe that MSE improves with training graph size and that GNN-based lower bounds outperform BFS-based lower bounds, even though the landmark embeddings are learned on much smaller graphs (see Figures~\ref{fig:exp3app} and \ref{fig:exp3app2}). This can be explained as random graphs can model real-world networks in certain scenarios, and networks with similar sparsity likely exhibit similar local structures which local message-passing in GNNs can learn with sufficient training.

\begin{table}[H]
\caption{Details on the largest connected component of selected benchmark networks.}
\centering
\resizebox{\textwidth}{!}{
\begin{tabular}{|c|c|c|c|c|} 
\hline
\# & Name & Category & \# of Nodes & \# of Edges \\
\hline
1 & Arxiv COND-MAT \citep{10.1145/1217299.1217301} & Collaboration Network & 21,364 & 91,315 \\
2 & Arxiv GR-QC \citep{10.1145/1217299.1217301} & Collaboration Network & 4,158 & 13,425 \\
3 & Arxiv HEP-PH \citep{10.1145/1217299.1217301} & Collaboration Network & 11,204 & 117,634 \\
4 & Arxiv HEP-TH \citep{10.1145/1217299.1217301} & Collaboration Network & 8,638 & 24,817 \\
5 & Oregon Autonomous System 1 \citep{10.1145/1081870.1081893} & Autonomous System & 11,174 & 23,409 \\
6 & Oregon Autonomous System 2 \citep{10.1145/1081870.1081893} & Autonomous System & 11,461 & 32,730 \\
7 & GEMSEC Athletes \citep{rozemberczki2019gemsec} & Social Network & 13,866 & 86,858 \\
8 & GEMSEC Public Figures \citep{rozemberczki2019gemsec} & Social Network & 11,565 & 67,114 \\
9 & GEMSEC Politicians \citep{rozemberczki2019gemsec} & Social Network & 5,908 & 41,729 \\
10 & GEMSEC Companies \citep{rozemberczki2019gemsec} & Social Network & 14,113 & 52,310 \\
11 & GEMSEC TV Shows \citep{rozemberczki2019gemsec} & Social Network & 3,892 & 17,262 \\
12 & Twitch-EN \citep{rozemberczki2019multiscale} & Social Network & 7,126 & 35,324 \\
13 & Deezer Europe \citep{feather} & Social Network & 28,281 & 92,752 \\
14 & LastFM Asia \citep{feather} & Social Network & 7,624 & 27,806 \\
15 & Brightkite \citep{nr} & Social Network & 56,739 & 212,945\\
16 & ER-AVGDEG10-100K-L2 \citep{nr} & Labeled Network & 99,997 & 499,359 \\
\hline
\end{tabular}\label{tab}
}
\end{table}



\begin{thebibliography}{59}
\providecommand{\natexlab}[1]{#1}
\providecommand{\url}[1]{\texttt{#1}}
\expandafter\ifx\csname urlstyle\endcsname\relax
  \providecommand{\doi}[1]{doi: #1}\else
  \providecommand{\doi}{doi: \begingroup \urlstyle{rm}\Url}\fi

\bibitem[Akiba et~al.(2013)Akiba, Iwata, and Yoshida]{akiba2013fast}
Takuya Akiba, Yoichi Iwata, and Yuichi Yoshida.
\newblock Fast exact shortest-path distance queries on large networks by pruned landmark labeling.
\newblock In \emph{Proceedings of the 2013 ACM SIGMOD International Conference on Management of Data}, pages 349--360, 2013.

\bibitem[Akiba et~al.(2014)Akiba, Iwata, and Yoshida]{Akiba2014queries}
Takuya Akiba, Yoichi Iwata, and Yuichi Yoshida.
\newblock Dynamic and historical shortest-path distance queries on large evolving networks by pruned landmark labeling.
\newblock In \emph{Proceedings of the 23rd International Conference on World Wide Web}, WWW '14, page 237–248, New York, NY, USA, 2014. Association for Computing Machinery.
\newblock ISBN 9781450327442.
\newblock \doi{10.1145/2566486.2568007}.
\newblock URL \url{https://doi.org/10.1145/2566486.2568007}.

\bibitem[Athreya and Ney(1972)]{AN72}
Krishna~B. Athreya and Peter~E. Ney.
\newblock \emph{{Branching Processes}}.
\newblock Springer, Berlin, Heidelberg, 1972.

\bibitem[Awasthi et~al.(2022)Awasthi, Das, and Gollapudi]{Awasthi2021}
Pranjal Awasthi, Abhimanyu Das, and Sreenivas Gollapudi.
\newblock Beyond {GNNs}: An efficient architecture for graph problems.
\newblock In \emph{Proceedings of the AAAI Conference on Artificial Intelligence}, volume~36, pages 6019--6027, 2022.

\bibitem[Belkin and Niyogi(2003)]{belkin2003laplacian}
Mikhail Belkin and Partha Niyogi.
\newblock Laplacian eigenmaps for dimensionality reduction and data representation.
\newblock \emph{Neural computation}, 15\penalty0 (6):\penalty0 1373--1396, 2003.

\bibitem[Bordenave(2016)]{Bordenave}
Charles Bordenave.
\newblock \emph{{Lecture notes on random graphs and probabilistic combinatorial optimization}}.
\newblock Available at https://www.math.univ-toulouse.fr/~bordenave/coursRG.pdf, 2016.

\bibitem[Bourgain(1985)]{Bourgain85}
J.~Bourgain.
\newblock On {L}ipschitz embedding of finite metric spaces in {H}ilbert space.
\newblock \emph{Israel Journal of Mathematics}, 52\penalty0 (1):\penalty0 46--52, 1985.
\newblock \doi{10.1007/BF02776078}.
\newblock URL \url{https://doi.org/10.1007/BF02776078}.

\bibitem[Brunner(2021)]{brunner2021distance}
Dustin Brunner.
\newblock \emph{Distance Preserving Graph Embedding}.
\newblock PhD thesis, BS thesis, ETH Z{\"u}rich, Zurich, 2021.[Online]. Available: https://pub. tik~…, 2021.

\bibitem[Cao et~al.(2015)Cao, Lu, and Xu]{cao2015grarep}
Shaosheng Cao, Wei Lu, and Qiongkai Xu.
\newblock Grarep: Learning graph representations with global structural information.
\newblock In \emph{Proceedings of the 24th ACM International on Conference on Information and Knowledge Management (CIKM)}, pages 891--900. ACM, 2015.

\bibitem[Cormen et~al.(2009)Cormen, Leiserson, Rivest, and Stein]{cormen2009introduction}
Thomas~H. Cormen, Charles~E. Leiserson, Ronald~L. Rivest, and Clifford Stein.
\newblock \emph{Introduction to Algorithms}.
\newblock MIT Press, Cambridge, MA, 3rd edition, 2009.

\bibitem[Dubhashi and Panconesi(2009)]{dubhashi2009concentration}
Devdatt~P Dubhashi and Alessandro Panconesi.
\newblock \emph{Concentration of measure for the analysis of randomized algorithms}.
\newblock Cambridge University Press, 2009.

\bibitem[Dudzik and Veli{\v{c}}kovi{\'c}(2022)]{dudzik2022graph}
Andrew~J Dudzik and Petar Veli{\v{c}}kovi{\'c}.
\newblock Graph neural networks are dynamic programmers.
\newblock \emph{Advances in neural information processing systems}, 35:\penalty0 20635--20647, 2022.

\bibitem[Fey and Lenssen(2019)]{fey2019fast}
Matthias Fey and Jan~Eric Lenssen.
\newblock Fast graph representation learning with {PyTorch Geometric}.
\newblock \emph{arXiv preprint arXiv:1903.02428}, 2019.

\bibitem[Fredman and Willard(1990)]{fredman1990trans}
Michael~L Fredman and Dan~E Willard.
\newblock Trans-dichotomous algorithms for minimum spanning trees and shortest paths.
\newblock In \emph{Proceedings [1990] 31st Annual Symposium on Foundations of Computer Science}, pages 719--725. IEEE, 1990.

\bibitem[Gallo and Pallottino(1988)]{gallo1988shortest}
Giorgio Gallo and Stefano Pallottino.
\newblock Shortest path algorithms.
\newblock \emph{Annals of operations research}, 13\penalty0 (1):\penalty0 1--79, 1988.

\bibitem[Goldberg and Harrelson(2005)]{goldberg2005computing}
Andrew~V Goldberg and Chris Harrelson.
\newblock Computing the shortest path: A search meets graph theory.
\newblock In \emph{SODA}, volume~5, pages 156--165, 2005.

\bibitem[Goyal and Ferrara(2018)]{goyal2018graph}
Palash Goyal and Emilio Ferrara.
\newblock Graph embedding techniques, applications, and performance: A survey.
\newblock In \emph{Knowledge-Based Systems}, volume 151, pages 78--94. Elsevier, 2018.

\bibitem[Grover and Leskovec(2016)]{grover2016node2vec}
Aditya Grover and Jure Leskovec.
\newblock node2vec: Scalable feature learning for networks.
\newblock In \emph{Proceedings of the 22nd ACM SIGKDD International Conference on Knowledge Discovery and Data Mining (KDD)}, pages 855--864. ACM, 2016.

\bibitem[Gubichev et~al.(2010)Gubichev, Bedathur, Seufert, and Weikum]{Gubichev2010}
Andrey Gubichev, Srikanta Bedathur, Stephan Seufert, and Gerhard Weikum.
\newblock Fast and accurate estimation of shortest paths in large graphs.
\newblock In \emph{Proceedings of the 19th ACM International Conference on Information and Knowledge Management}, CIKM '10, page 499–508, New York, NY, USA, 2010. Association for Computing Machinery.
\newblock ISBN 9781450300995.
\newblock \doi{10.1145/1871437.1871503}.
\newblock URL \url{https://doi.org/10.1145/1871437.1871503}.

\bibitem[Hamilton et~al.(2017{\natexlab{a}})Hamilton, Ying, and Leskovec]{NIPS2017_5dd9db5e}
Will Hamilton, Zhitao Ying, and Jure Leskovec.
\newblock Inductive representation learning on large graphs.
\newblock In I.~Guyon, U.~Von Luxburg, S.~Bengio, H.~Wallach, R.~Fergus, S.~Vishwanathan, and R.~Garnett, editors, \emph{Advances in Neural Information Processing Systems}, volume~30. Curran Associates, Inc., 2017{\natexlab{a}}.

\bibitem[Hamilton et~al.(2017{\natexlab{b}})Hamilton, Ying, and Leskovec]{hamilton2017representation}
William~L Hamilton, Rex Ying, and Jure Leskovec.
\newblock Representation learning on graphs: Methods and applications.
\newblock In \emph{IEEE Data Engineering Bulletin}, volume~40, pages 52--74, 2017{\natexlab{b}}.

\bibitem[{\swap{Hofstad}{van der }}(2017)]{RGCN1}
Remco {\swap{Hofstad}{van der }}.
\newblock \emph{{Random Graphs and Complex Networks}}, volume~I.
\newblock Cambridge University Press, Cambridge, 2017.
\newblock \doi{10.1017/9781316779422}.
\newblock URL \url{http://www.win.tue.nl/{~}rhofstad/NotesRGCN.pdf}.

\bibitem[{\swap{Hofstad}{van der }}(2024)]{RGCN2}
Remco {\swap{Hofstad}{van der }}.
\newblock \emph{{Random Graphs and Complex Networks}}, volume~II.
\newblock Cambridge University Press, Cambridge, 2024.

\bibitem[Janson et~al.(2000)Janson, Luczak, and Rucinski]{JLR00}
Svante Janson, Tomasz Luczak, and Andrzej Rucinski.
\newblock \emph{Random Graphs}.
\newblock John Wiley \& Sons, 2000.

\bibitem[Jiang et~al.(2021)Jiang, Xu, Yin, Zhao, and Gupta]{Jiang2021Tripoline}
Xiaolin Jiang, Chengshuo Xu, Xizhe Yin, Zhijia Zhao, and Rajiv Gupta.
\newblock Tripoline: generalized incremental graph processing via graph triangle inequality.
\newblock In \emph{Proceedings of the Sixteenth European Conference on Computer Systems}, EuroSys '21, page 17–32, New York, NY, USA, 2021. Association for Computing Machinery.
\newblock ISBN 9781450383349.
\newblock \doi{10.1145/3447786.3456226}.
\newblock URL \url{https://doi.org/10.1145/3447786.3456226}.

\bibitem[Johnson and Lindenstrauss(1984)]{JL84}
W.~B. Johnson and J.~Lindenstrauss.
\newblock Extensions of lipschitz mappings into a hilbert space.
\newblock In \emph{Conference in Modern Analysis and Probability (New Haven, Conn., 1982)}, volume~26 of \emph{Contemporary Mathematics}, pages 189--206. American Mathematical Society, Providence, RI, 1984.

\bibitem[Karger et~al.(1993)Karger, Koller, and Phillips]{karger1993finding}
David~R Karger, Daphne Koller, and Steven~J Phillips.
\newblock Finding the hidden path: Time bounds for all-pairs shortest paths.
\newblock \emph{SIAM Journal on Computing}, 22\penalty0 (6):\penalty0 1199--1217, 1993.

\bibitem[Kipf and Welling(2017)]{kipf17-classifgcnn}
T.~N. Kipf and M.~Welling.
\newblock Semi-supervised classification with graph convolutional networks.
\newblock In \emph{5th Int. Conf. Learning Representations}, Toulon, France, 24-26 Apr. 2017. Assoc. Comput. Linguistics.

\bibitem[Leskovec et~al.(2005)Leskovec, Kleinberg, and Faloutsos]{10.1145/1081870.1081893}
Jure Leskovec, Jon Kleinberg, and Christos Faloutsos.
\newblock Graphs over time: densification laws, shrinking diameters and possible explanations.
\newblock In \emph{Proceedings of the Eleventh ACM SIGKDD International Conference on Knowledge Discovery in Data Mining}, KDD '05, page 177–187, New York, NY, USA, 2005. Association for Computing Machinery.
\newblock ISBN 159593135X.
\newblock \doi{10.1145/1081870.1081893}.
\newblock URL \url{https://doi.org/10.1145/1081870.1081893}.

\bibitem[Leskovec et~al.(2007)Leskovec, Kleinberg, and Faloutsos]{10.1145/1217299.1217301}
Jure Leskovec, Jon Kleinberg, and Christos Faloutsos.
\newblock Graph evolution: Densification and shrinking diameters.
\newblock \emph{ACM Trans. Knowl. Discov. Data}, 1\penalty0 (1):\penalty0 2–es, mar 2007.
\newblock ISSN 1556-4681.
\newblock \doi{10.1145/1217299.1217301}.
\newblock URL \url{https://doi.org/10.1145/1217299.1217301}.

\bibitem[Linial et~al.(1995)Linial, London, and Rabinovich]{LLR95}
N.~Linial, E.~London, and Y.~Rabinovich.
\newblock The geometry of graphs and some of its algorithmic applications.
\newblock \emph{Combinatorica}, 15\penalty0 (2):\penalty0 215--245, 1995.

\bibitem[Loukas(2020)]{Loukas2020}
Andreas Loukas.
\newblock What graph neural networks cannot learn: depth vs width.
\newblock In \emph{International Conference on Learning Representations}, 2020.
\newblock URL \url{https://openreview.net/forum?id=B1l2bp4YwS}.

\bibitem[Matou{\v{s}}ek(1996)]{Mat96}
J.~Matou{\v{s}}ek.
\newblock On the distortion required for embedding finite metric spaces into normed spaces.
\newblock \emph{Israel Journal of Mathematics}, 93:\penalty0 333--344, 1996.

\bibitem[Meng et~al.(2015)Meng, Kamara, Nissim, and Kollios]{Meng2015GRECS}
Xianrui Meng, Seny Kamara, Kobbi Nissim, and George Kollios.
\newblock Grecs: Graph encryption for approximate shortest distance queries.
\newblock In \emph{Proceedings of the 22nd ACM SIGSAC Conference on Computer and Communications Security}, CCS '15, page 504–517, New York, NY, USA, 2015. Association for Computing Machinery.
\newblock ISBN 9781450338325.
\newblock \doi{10.1145/2810103.2813672}.
\newblock URL \url{https://doi.org/10.1145/2810103.2813672}.

\bibitem[Naor(2016)]{Naor16}
Assaf Naor.
\newblock A spectral gap precludes low-dimensional embeddings.
\newblock \emph{ArXiv}, abs/1611.08861, 2016.
\newblock URL \url{https://api.semanticscholar.org/CorpusID:2112685}.

\bibitem[Naor(2021)]{Naor21}
Assaf Naor.
\newblock An average john theorem.
\newblock \emph{Geometry \& Topology}, 25\penalty0 (4):\penalty0 1631--1717, 2021.
\newblock \doi{10.2140/gt.2021.25.1631}.

\bibitem[Perozzi et~al.(2014)Perozzi, Al-Rfou, and Skiena]{perozzi2014deepwalk}
Bryan Perozzi, Rami Al-Rfou, and Steven Skiena.
\newblock Deepwalk: Online learning of social representations.
\newblock In \emph{Proceedings of the 20th ACM SIGKDD International Conference on Knowledge Discovery and Data Mining (KDD)}, pages 701--710. ACM, 2014.

\bibitem[Potamias et~al.(2009)Potamias, Bonchi, Castillo, and Gionis]{potamias2009fast}
Michalis Potamias, Francesco Bonchi, Carlos Castillo, and Aristides Gionis.
\newblock Fast shortest path distance estimation in large networks.
\newblock In \emph{Proceedings of the 18th ACM conference on Information and knowledge management}, pages 867--876, 2009.

\bibitem[Qi et~al.(2020)Qi, Wang, Zhang, and Zhao]{Qi2020ALB}
Jianzhong Qi, Wei Wang, Rui Zhang, and Zhuowei Zhao.
\newblock A learning based approach to predict shortest-path distances.
\newblock In \emph{International Conference on Extending Database Technology}, 2020.
\newblock URL \url{https://api.semanticscholar.org/CorpusID:214613433}.

\bibitem[Rizi et~al.(2018)Rizi, Schloetterer, and Granitzer]{rizi2018shortest}
Fatemeh~Salehi Rizi, Joerg Schloetterer, and Michael Granitzer.
\newblock Shortest path distance approximation using deep learning techniques.
\newblock In \emph{2018 IEEE/ACM International Conference on Advances in Social Networks Analysis and Mining (ASONAM)}, pages 1007--1014. IEEE, 2018.

\bibitem[Rossi and Ahmed(2015)]{nr}
Ryan~A. Rossi and Nesreen~K. Ahmed.
\newblock The network data repository with interactive graph analytics and visualization.
\newblock In \emph{AAAI}, 2015.
\newblock URL \url{https://networkrepository.com}.

\bibitem[Rozemberczki and Sarkar(2020)]{feather}
Benedek Rozemberczki and Rik Sarkar.
\newblock Characteristic functions on graphs: Birds of a feather, from statistical descriptors to parametric models.
\newblock In \emph{Proceedings of the 29th ACM International Conference on Information and Knowledge Management (CIKM '20)}, page 1325–1334. ACM, 2020.

\bibitem[Rozemberczki et~al.(2019{\natexlab{a}})Rozemberczki, Allen, and Sarkar]{rozemberczki2019multiscale}
Benedek Rozemberczki, Carl Allen, and Rik Sarkar.
\newblock Multi-scale attributed node embedding, 2019{\natexlab{a}}.

\bibitem[Rozemberczki et~al.(2019{\natexlab{b}})Rozemberczki, Davies, Sarkar, and Sutton]{rozemberczki2019gemsec}
Benedek Rozemberczki, Ryan Davies, Rik Sarkar, and Charles Sutton.
\newblock Gemsec: Graph embedding with self clustering.
\newblock In \emph{Proceedings of the 2019 IEEE/ACM International Conference on Advances in Social Networks Analysis and Mining 2019}, pages 65--72. ACM, 2019{\natexlab{b}}.

\bibitem[Ruiz et~al.(2020)Ruiz, Chamon, and Ribeiro]{ruiz20-transf}
L.~Ruiz, L.~F.~O. Chamon, and A.~Ribeiro.
\newblock Graphon neural networks and the transferability of graph neural networks.
\newblock In \emph{34th Neural Inform. Process. Syst.}, Vancouver, BC (Virtual), 6-12 Dec. 2020. NeurIPS Foundation.

\bibitem[Ruiz et~al.(2023)Ruiz, Chamon, and Ribeiro]{ruiz2021transferability}
L.~Ruiz, L.~F.~O. Chamon, and A.~Ribeiro.
\newblock Transferability properties of graph neural networks.
\newblock \emph{IEEE Transactions on Signal Processing}, 2023.

\bibitem[Sarma et~al.(2010)Sarma, Gollapudi, Najork, and Panigrahy]{Sarma2010ASD}
Atish~Das Sarma, Sreenivas Gollapudi, Marc Najork, and Rina Panigrahy.
\newblock A sketch-based distance oracle for web-scale graphs.
\newblock In \emph{Web Search and Data Mining}, 2010.
\newblock URL \url{https://api.semanticscholar.org/CorpusID:17378629}.

\bibitem[Sarma et~al.(2012)Sarma, Holzer, Kor, Korman, Nanongkai, Pandurangan, Peleg, and Wattenhofer]{Sarma2012}
Atish~Das Sarma, Stephan Holzer, Liah Kor, Amos Korman, Danupon Nanongkai, Gopal Pandurangan, David Peleg, and Roger Wattenhofer.
\newblock Distributed verification and hardness of distributed approximation.
\newblock \emph{SIAM Journal on Computing}, 41\penalty0 (5):\penalty0 1235--1265, 2012.
\newblock \doi{10.1137/11085178X}.
\newblock URL \url{https://doi.org/10.1137/11085178X}.

\bibitem[Schrijver(2012)]{schrijver2012history}
Alexander Schrijver.
\newblock On the history of the shortest path problem.
\newblock \emph{Documenta Mathematica}, 17\penalty0 (1):\penalty0 155--167, 2012.

\bibitem[Sidiropoulos et~al.(2019)Sidiropoulos, Badoiu, Dhamdhere, Gupta, Indyk, Rabinovich, Racke, and Ravi]{Sid19}
Anastasios Sidiropoulos, Mihai Badoiu, Kedar Dhamdhere, Anupam Gupta, Piotr Indyk, Yuri Rabinovich, Harald Racke, and R.~Ravi.
\newblock Approximation algorithms for low-distortion embeddings into low-dimensional spaces.
\newblock \emph{SIAM Journal on Discrete Mathematics}, 33\penalty0 (1):\penalty0 454--473, 2019.
\newblock \doi{10.1137/17M1113527}.
\newblock URL \url{https://doi.org/10.1137/17M1113527}.

\bibitem[Sommer(2014)]{Sommer2014queries}
Christian Sommer.
\newblock Shortest-path queries in static networks.
\newblock \emph{ACM Comput. Surv.}, 46\penalty0 (4), March 2014.
\newblock ISSN 0360-0300.
\newblock \doi{10.1145/2530531}.
\newblock URL \url{https://doi.org/10.1145/2530531}.

\bibitem[Tang et~al.(2019)Tang, Lu, and Zhu]{tang2019cauchy}
Jian Tang, Yifan Lu, and Xia Zhu.
\newblock Cauchy graph embedding.
\newblock \emph{Journal of Machine Learning Research}, 20\penalty0 (1):\penalty0 1--27, 2019.

\bibitem[Tanny(1977)]{Tanny77}
David Tanny.
\newblock Limit theorems for branching processes in a random environment.
\newblock \emph{The Annals of Probability}, 5\penalty0 (1):\penalty0 100 -- 116, 1977.
\newblock \doi{10.1214/aop/1176995894}.
\newblock URL \url{https://doi.org/10.1214/aop/1176995894}.

\bibitem[Tretyakov et~al.(2011)Tretyakov, Armas-Cervantes, Garc{\'\i}a-Ba{\~n}uelos, Vilo, and Dumas]{tretyakov2011fast}
Konstantin Tretyakov, Abel Armas-Cervantes, Luciano Garc{\'\i}a-Ba{\~n}uelos, Jaak Vilo, and Marlon Dumas.
\newblock Fast fully dynamic landmark-based estimation of shortest path distances in very large graphs.
\newblock In \emph{Proceedings of the 20th ACM international conference on Information and knowledge management}, pages 1785--1794, 2011.

\bibitem[Tsitsulin et~al.(2018)Tsitsulin, Mottin, Karras, and M{\"u}ller]{tsitsulin2018just}
Anton Tsitsulin, Davide Mottin, Panagiotis Karras, and Emmanuel M{\"u}ller.
\newblock Just walk the graph: Learning node embeddings via meta paths.
\newblock In \emph{Proceedings of the 2018 IEEE International Conference on Data Mining (ICDM)}, pages 1306--1311. IEEE, 2018.

\bibitem[Veli{\v{c}}kovi{\'{c}} et~al.(2018)Veli{\v{c}}kovi{\'{c}}, Cucurull, Casanova, Romero, Li{\`{o}}, and Bengio]{Velickovic18-GraphAttentionNetworks}
P.~Veli{\v{c}}kovi{\'{c}}, G.~Cucurull, A.~Casanova, A.~Romero, P.~Li{\`{o}}, and Y.~Bengio.
\newblock Graph attention networks.
\newblock In \emph{Int. Conf. Learning Representations 2018}, pages 1--12, Vancouver, BC, 30 Apr.-3 May 2018. Assoc. Comput. Linguistics.

\bibitem[Xu et~al.(2019{\natexlab{a}})Xu, Hu, Leskovec, and Jegelka]{xu2018how}
K.~Xu, W.~Hu, J.~Leskovec, and S.~Jegelka.
\newblock How powerful are graph neural networks?
\newblock In \emph{7th Int. Conf. Learning Representations}, pages 1--17, New Orleans, LA, 6-9 May 2019{\natexlab{a}}. Assoc. Comput. Linguistics.

\bibitem[Xu et~al.(2019{\natexlab{b}})Xu, Li, Zhang, Du, Kawarabayashi, and Jegelka]{xu2019can}
Keyulu Xu, Jingling Li, Mozhi Zhang, Simon~S Du, Ken-ichi Kawarabayashi, and Stefanie Jegelka.
\newblock What can neural networks reason about?
\newblock \emph{arXiv preprint arXiv:1905.13211}, 2019{\natexlab{b}}.

\bibitem[Zhang et~al.(2021)Zhang, Zhang, Feng, Wang, and Zhang]{zhang2021prone}
Wen Zhang, Yuxuan Zhang, Yansong Feng, Zheng Wang, and Jin Zhang.
\newblock Prone: A scalable graph embedding method with local proximity preservation.
\newblock \emph{IEEE Transactions on Knowledge and Data Engineering}, 33\penalty0 (6):\penalty0 2500--2513, 2021.

\end{thebibliography}
\end{document}